%% file: main.tex
\documentclass[letterpaper, 10 pt, conference]{ieeeconf}  % Comment this line out if you need a4paper
\IEEEoverridecommandlockouts                              % This command is only needed if
\usepackage{epsfig} % for postscript graphics files
\usepackage{mathptmx} % assumes new font selection scheme installed
\usepackage{times} % assumes new font selection scheme installed
\usepackage{amsmath} % assumes amsmath package installed
\usepackage{amssymb}  % assumes amsmath package installed
\newcommand{\debug}{0}
\input{lib/daimler_color.tex}   %add by rheinzl
\input{lib/settings.tex}   %add by rheinzl
\title{\LARGE \bf
CNN-based Lidar Point Cloud De-Noising in Adverse Weather
}
\author{Robin Heinzler$^{1,2}$, Florian Piewak$^{1}$, Philipp Schindler$^{1}$ and Wilhelm Stork$^{2}$% <-this % stops a space
\thanks{$^{1}$Daimler AG, Benz-Str., 71063 Sindelfingen, Germany
        {\tt\small [firstname.lastname]@daimler.com}}%
\thanks{$^{2}$Institute for Information Processing Technology (ITIV), Karlsruhe Institute of Technology (KIT), Germany
		{\tt\small [firstname.lastname]@kit.edu}}%
}
\begin{document}%
\maketitle
\thispagestyle{empty}
\pagestyle{empty}
%
% copyright IEEE e.g. for arxiv
\input{tex/ieee_copyright.tex}

%
\input{tex/abstract} %
\section{INTRODUCTION}
\input{tex/introduction}
\section{RELATED WORK}
\input{tex/related_work} %
\section{METHOD}
\input{tex/method} %
\section{DATASET}
\input{tex/dataset} %
\section{EXPERIMENTS} %
\input{tex/experiments} %
\section{CONCLUSION}
\input{tex/conclusion}
\input{tex/acknowledgment}
%
%\addtolength{\textheight}{-15cm}
\bibliographystyle{IEEEtran}
\bibliography{bib/PhD-ICRA2020}
\end{document}

%% file: lib/daimler_color.tex
\usepackage{color}
\usepackage[rgb,pdftex,dvipsnames, table]{xcolor}    %rh
\definecolor{daiCoolGrey} {RGB}{230,230,230}
\definecolor{daiCoolGrey1}{RGB}{68,68,68}
\definecolor{daiCoolGrey2}{RGB}{112,112,112}
\definecolor{daiCoolGrey3}{RGB}{158,158,158}
\definecolor{daiCoolGrey4}{RGB}{200,200,200}
\definecolor{daiPetrol} {RGB}{0,103,127}
\definecolor{daiPetrol1}{RGB}{0,51,60}
\definecolor{daiPetrol2}{RGB}{0,67,85}
\definecolor{daiPetrol3}{RGB}{0,86,106}
\definecolor{daiPetrol4}{RGB}{0,122,147}
\definecolor{daiPetrol5}{RGB}{80,151,171}
\definecolor{daiPetrol6}{RGB}{121,174,191}
\definecolor{daiPetrol7}{RGB}{166,202,216}
\definecolor{daiDeepRed} {RGB}{114,23,12}
\definecolor{daiDeepRed1}{RGB}{68,14,7}
\definecolor{daiDeepRed2}{RGB}{90,19,10}
\definecolor{daiDeepRed3}{RGB}{159,25,36}
\definecolor{daiDeepRed4}{RGB}{255,0,0}
\definecolor{daiDeepRed5}{RGB}{140,70,60}
\definecolor{daiDeepRed6}{RGB}{170,115,110}
\definecolor{daiDeepRed7}{RGB}{200,160,160}
\definecolor{daiDeepRed8}{RGB}{230,210,210}
\definecolor{daiOrange} {RGB}{230,145,35}
\definecolor{daiOrange1}{RGB}{235,165,80}
\definecolor{daiOrange2}{RGB}{240,190,125}
\definecolor{daiOrange3}{RGB}{245,210,170}
\definecolor{daiOrange4}{RGB}{250,230,210}

\definecolor{daiGreen} {RGB}{110,160,70}
\definecolor{daiGreen1}{RGB}{140,180,110}
\definecolor{daiGreen2}{RGB}{170,200,145}
\definecolor{daiGreen3}{RGB}{200,200,180}
\definecolor{daiGreen4}{RGB}{225,235,220}

\definecolor{colorFog}{RGB}{115,0,230}
\definecolor{colorRain}{RGB}{0,153,153}
\definecolor{colorClear}{RGB}{0,0,0}
\definecolor{colorValid}{named}{daiCoolGrey3}
\definecolor{colorInput}{named}{daiDeepRed3} %daiDeepRed
\definecolor{colorDenoised}{named}{daiGreen}

\definecolor{color0}{named}{daiPetrol}
\definecolor{color1}{named}{daiDeepRed}
\definecolor{color2}{named}{daiOrange}
\definecolor{color3}{named}{daiGreen}
\definecolor{color4}{named}{daiCoolGrey2}

\definecolor{boxplot_whisker}{named}{black}
\definecolor{boxplot_median}{named}{daiPetrol}
\definecolor{boxplot_frame}{named}{black}
\definecolor{boxplot_fill}{named}{white}
\definecolor{boxplot_outlier}{named}{daiDeepRed}

\definecolor{maxcolor}{named}{blue}

%% file: lib/settings.tex
\usepackage{cases} %
\usepackage{makeidx}
\usepackage{graphicx}
\usepackage[caption=false, font=footnotesize]{subfig}
\usepackage{amsmath,amssymb} % define this before the line numbering.
\usepackage{color}
\usepackage{tikz}
\usepackage{cite}           %rh
\usepackage{verbatim} 		%rh
\usepackage{import}			%rh
\graphicspath{{img/}}	%rh
\usepackage{multirow}       %rh
\usepackage{booktabs}       %rh
\usepackage{lipsum}                     % Dummytext
\usepackage{xargs}                      % Use more than one optional parameter in a new commands
\usepackage{standalone}
\usepackage{tabularx}       %rh
\usepackage[ruled,noend]{algorithm2e} % rh for algorithms http://tug.ctan.org/macros/latex/contrib/algorithm2e/doc/algorithm2e.pdf

\SetAlFnt{\footnotesize}
\SetAlCapFnt{\footnotesize}
\SetAlCapNameFnt{\footnotesize}
\SetKwComment{Comment}{$\triangleright$\ }{}

\usepackage{csvsimple}
\usepackage[nomessages]{fp}% http://ctan.org/pkg/fp # rh for calculation in text
\usepackage[group-separator={,}]{siunitx}
\usepackage{etoolbox}
\robustify\bfseries
\usepackage{hyperref}
\hypersetup{
	colorlinks=true,
	linkcolor=black,
	filecolor=magenta,      
	urlcolor=black,
	citecolor=green,
}
\usepackage{url}            % simple URL typesetting

\usetikzlibrary{shapes,shapes.multipart} % Required for the trapezoid shape

\usepackage{pgfplots}
\usepackage{pgfplotstable} 	%add by rheinzl
\usepackage{datatool}
\definecolor{MyRed}{RGB}{0,0,0}
\definecolor{MyGreen}{RGB}{0,0,0}
\definecolor{MyBlue}{RGB}{0,0,0}

\newcolumntype{P}[1]{>{\centering\arraybackslash}p{#1}}

\FPeval{\mIoU}{clip(81.67)}%
\FPeval{\mIoUDROR}{clip(34.15)}%

\ifnum\debug=2

	\usepackage[colorinlistoftodos,prependcaption,textsize=footnotesize]{todonotes}
\else
	\ifnum\debug=1

		\usepackage[colorinlistoftodos,prependcaption,textsize=footnotesize,disable]{todonotes}
	\else

		\usepackage[colorinlistoftodos,prependcaption,textsize=footnotesize,disable]{todonotes}
	\fi
\fi

\newcommandx{\unsure}[2][1=]{\todo[linecolor=red,backgroundcolor=red!25,bordercolor=red,#1]{#2}}
\newcommandx{\change}[2][1=]{\todo[linecolor=blue,backgroundcolor=blue!25,bordercolor=blue,#1]{#2}}
\newcommandx{\info}[2][1=]{\todo[linecolor=OliveGreen,backgroundcolor=OliveGreen!25,bordercolor=OliveGreen,#1]{\scriptsize{#2}}}
\newcommandx{\improvement}[2][1=]{\todo[linecolor=Plum,backgroundcolor=Plum!25,bordercolor=Plum,#1]{#2}}
\newcommandx{\TODO}[2][1=]{\todo[linecolor=red,backgroundcolor=red!25,bordercolor=red,#1]{#2}}
\newcommandx{\thiswillnotshow}[2][1=]{\todo[disable,#1]{#2}}
\usetikzlibrary{calc}
\newcommand{\positiontextbox}[4][]{%
  \begin{tikzpicture}[remember picture,overlay]
    \node[inner sep=3pt,right,draw,line width=1pt,#1, text width=2.2\linewidth, align=center] at ($(current page.south west) + (#2,-#3)$)%
    {\baselineskip=5pt\scriptsize{#4\par}};
  \end{tikzpicture}%
}

%% file: tex/ieee_copyright.tex
\positiontextbox{1.1cm}{-1.25cm}{ %
%\textsuperscript{\textcopyright} 2019 IEEE. Personal use of this material is permitted. Permission from IEEE must be obtained for all other uses, in any current or future media, including reprinting/republishing this material for advertising or promotional purposes, creating new collective works, for resale or redistribution to servers or lists, or reuse of any copyrighted component of this work in other works. %
\textsuperscript{\textcopyright}\,2019\,IEEE.\,Personal\,use\,of\,this\,material\,is\,permitted.\,Permission\,from\,IEEE\,must\,be\,obtained\,for\,all\,other\,uses,\,in\,any\,current\,or\,future\,media,\,including\,reprinting/republishing\,this\,material for\,advertising\,or\,promotional\,purposes,\,creating\,new\,collective\,works,\,for\,resale\,or\,redistribution\,to\,servers\,or\,lists,\,or\,reuse\,of\,any\,copyrighted\,component\,of\,this\,work\,in\,other\,works.\,%
%(This paper has been submitted to the IEEE for publication.)
%Copyright may be transferred without notice, after which this edition may no longer be available.
}
%
%\positiontextbox{1.505cm}{-1.4cm}{This paper has been submitted to the IEEE for publication. Copyright may be transferred without notice, after which this edition may no longer be available.}

%% file: tex/abstract.tex
\begin{abstract}	
Lidar sensors are frequently used in environment perception for autonomous vehicles and mobile robotics to complement camera, radar, and ultrasonic sensors. 
Adverse weather conditions are significantly impacting the performance of lidar-based scene understanding by causing undesired measurement points that in turn effect missing detections and false positives. 
%Thus filtering of weather influences is extremely important for further processing algorithms such as object recognition.
%Otherwise, weather effects can lead to misclassifications and false positives. 
%In extreme cases, this can lead to no free space being detected in heavy rain or dense fog.
In heavy rain or dense fog, water drops could be misinterpreted as objects in front of the vehicle which brings a mobile robot to a full stop. 

%this can lead to no free space being detected in heavy rain or dense fog, as water drops are detected as objects directly in front of the vehicle.

In this paper, 
we present the first \textit{CNN}-based approach to understand and filter out such adverse weather effects in point cloud data. Using a 
large data set obtained in controlled weather environments, we demonstrate a significant performance improvement of our method over 
state-of-the-art involving geometric filtering. Data is available at {\normalfont\textit{
\href{https://github.com/rheinzler/PointCloudDeNoising}{https://github.com/rheinzler/PointCloudDeNoising}}}.
%In our work we have developed a novel approach for filtering weather influences such as rain or fog based on convolutional neural networks with a mean intersection over union (IoU) of \textbf{$\mIoU\,\%$} for point-wise classifications. 
%In contrast, geometrically filtering algorithms tend to reduce the overall detection performance of sensors as objects in farther distances are filtered as well due to the low density of reflections. \DRAFT{Furthermore, geometrical approaches based on the spatial vicinity are not suited to filter dense scatter clouds as the mean IoU of $\mIoUDROR\,\%$ is indicating.}
%As conventional filter algorithms, using spatial vicinity as a basis, are restricting the range of sensors by increasingly filtering points at greater distances. 
%To validate the algorithm, we recorded a large data set in controlled environments under different weather conditions and automatically labeled it based on the reference measurements. 
\end{abstract}
%are important steps to make autonomous driving in bad weather safer 

%% file: tex/introduction.tex
\info{Lidar Sensors are the enabler, but not in adverse weather}
Given that lidar sensors are key for autonomous driving and robotics applications, they are currently being developed
by numerous companies in a wide variety of designs. Nevertheless, lidar technology is heavily challenged 
% conceptually not suitable for being used 
in adverse weather as the range measurements are highly impaired by fog, dust, snow, rain,
pollution, and smog \cite{Bijelic2018,Heinzler2019,Phillips2017,Charron2018,Kutila2018,Djuricic2013}. Such conditions
cause erroneous point measurements in the point cloud data which arise from the reception of back-scattered light
from water drops (e.g rain or fog) or arbitrary  particles in the air (e.g. smog or dust).

For environment perception algorithms, these points are undesirable noise which needs to be specifically addressed in
order to not restrain the scene understanding performance. This is particularly relevant for algorithms that make
direct use of the low-level geometry of a measured point cloud, e.g. the Stixel algorithm~\cite{Piewak2019}, where
noisy input data inevitably results in noisy Stixel output data. \textit{CNN}-based lidar perception algorithms~\TODO{cite
some scene labeling paper here} might be better able to cope with such issues given their learning capacity thereby
reducing the need for an explicit handling of noisy measurements. Still, most lidar perception algorithms involve
more classical bottom-up approaches for tasks such as object detection since they usually are implemented on the
lidar sensors themselves with very limited computational resources. This has sparked a large body of research on
algorithms to detect and handle noisy point cloud measurements in a pre-processing step before applying perception
algorithms.

\input{img/fig_predict_results_denoising/denoise_results_dynamic_header.tex}
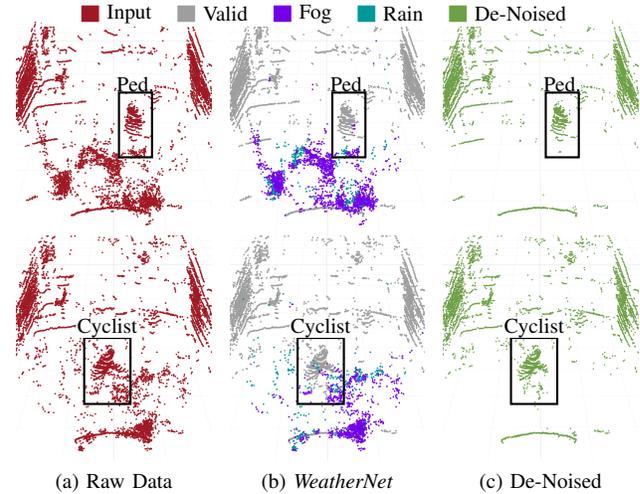
\begin{figure}
	\centering
	\input{img/fig_predict_results_denoising/denoise_results_dynamic.tex}
	\caption{De-noising results shown on snapshots of two dynamic scenes in dense fog at $20$ and $30\,$m visibility. Fig. (a) shows the raw 
	point-cloud data (\textcolor{colorInput}{red}), (b) the point-wise weather segmentation by \textit{WeatherNet} 
	(\textcolor{colorFog}{fog}, \textcolor{colorRain}{rain}, \textcolor{colorValid}{valid}) and (c) the de-noised result 
	(\textcolor{colorDenoised}{green}) with the remaining valid points, where reflections from fog and rain are removed. The pedestrian and 
	cyclist, which are barely recognizable in the scene, are highlighted in a black box.
	}\vspace{-0.5cm}%%
	\label{fig:denoise_results_Dynamic-G2-Fog20} %
\end{figure} %

To that extent, a large quantity of 2D image anti-aliasing algorithms have been developed that focus on smoothing
noisy surface points resulting from marginal sensor errors \cite{Tomasi1998,Ronnback2008,Li2014a,Jenke2006,Chen2012,
Shen2013,Le2014}.
\info{challenge of discard noise of rain and snow, but preserve sparse points in large distances}
De-noising algorithms in 3D space are often based on spatial features to discard noise points caused by rain or snow \cite{Rusu2011,Charron2018}.
As these techniques are discarding points based on the absence of points in their vicinity, smaller objects at medium
to large distances might be falsely suppressed and marked as noise.
In addition, the recordings in Fig. \ref{fig:denoise_results_Dynamic-G2-Fog20} indicate that modern lidar sensors, e.g. the
Velodyne VLP32C, do not necessarily perceive drops of water from fog or rain as a single point, but often as
multi-point clutter in the near to mid range which significantly reduces the applicability of filtering based on
spatial vicinity only.

Experimentally validating such filtering algorithms in real-world scenarios under adverse weather conditions is very
challenging due to the lack of proper ground truth. We address this by proposing an evaluation based on data recorded
in controlled environments where we could obtain a large data set under various environmental conditions with
point-wise ground truth annotations for different classes of weather, e.g.~\textit{clear, rain,} or \textit{fog}. In addition,
we employ a data augmentation approach to emulate adverse weather effects on real-world data that has been previously
obtained in good weather conditions \cite{Bijelic2019}.

Our main contributions are as follows:
\begin{itemize}
\item {The first \textit{CNN}-based approach to lidar point cloud de-noising with a significant performance boost
 over previous state-of-the-art while being very efficient at the same time.}
\item {A data augmentation approach for adding realistic weather effects to lidar point cloud data.}
\item {A quantitative and qualitative point-level evaluation of de-noising algorithms in controlled environments
 under different weather conditions.}
\end{itemize}

%% file: img/fig_predict_results_denoising/denoise_results_dynamic_header.tex
\newcommand{\resultFolder}{./fig_predict_results_denoising/}
\renewcommand{\resultFolder}{./csv/resubmit_03/fig_predict_results_denoise/}
%-----------------------------------------------------------------------------------------------------------------------
%------------------Models
%\newcommand{\weatherNetInitSub}{semseg_di_32_400_aug_both_sprinkles_full_LiLaBlockDil_1e3_2k5_ckp_40k_freeze/}
\newcommand{\weatherNetInitSub}{./csv/resubmit_03/fig_predict_results_denoise_initital_WeatherNet/semseg_di_32_400_aug_both_sprinkles_full_LiLaBlockDil_1e3_2k5_ckp_40k_freeze/}
\newcommand{\weatherNet}{resubmission_02/exp03_WeatherNet/}
\renewcommand{\weatherNet}{exp03_WeatherNet/exp03_WeatherNet_}
\newcommand{\weatherNetExpOne}{resubmission_02/exp01_WeatherNet/}
\newcommand{\weatherNetExpTwo}{resubmission_02/exp02_WeatherNet/}
\renewcommand{\weatherNetExpTwo}{exp02_WeatherNet/exp02_WeatherNet_}
\newcommand{\weatherNetinUse}{\weatherNet}
\newcommand{\rangeNet}{exp03_RangeNet53_v2/exp03_RangeNet53_v2_}
%------------------DROR
\newcommand{\drorPedC}{DROR_baseline/2018-11-26_153228_Dynamic-G2-Fog20_LidarImage_000000285}
\newcommand{\drorCycC}{DROR_baseline/2018-11-26_155631_Dynamic-G6-Fog30_LidarImage_000000175}
%-----------------------------------------------------------------------------------------------------------------------
%------------------pedestrian and cylcist
\newcommand{\pathPedA}{pedestrian_2018-11-26_153228_Dynamic-G2-Fog20_LidarImage_000000267}%
\newcommand{\pathPedB}{pedestrian_2018-11-26_153228_Dynamic-G2-Fog20_LidarImage_000000280}%
\newcommand{\pathPedC}{pedestrian_2018-11-26_153228_Dynamic-G2-Fog20_LidarImage_000000285}%
\newcommand{\pathCycA}{cyclist_2018-11-26_153359_Dynamic-G4-Fog20_LidarImage_000000237}%
\newcommand{\pathCycB}{cyclist_2018-11-26_155631_Dynamic-G6-Fog30_LidarImage_000000187}%
\newcommand{\pathCycC}{cyclist_2018-11-26_155631_Dynamic-G6-Fog30_LidarImage_000000175}%
%-----------------------------------------------------------------------------------------------------------------------
%------------------Road
\newcommand{\pathRoadA}{road_2018-01-17_143024_LidarImage_000000919}%
\newcommand{\pathRoadB}{road_2018-02-12_142735_LidarImage_000000284}%

\tikzstyle{txt}=[fill=none,inner sep=1pt, text centered,minimum height=1.0em, font=\footnotesize]
\tikzstyle{box}=[fill=white,inner sep=1pt,text width=2.45cm, text centered,minimum height=1.0em]
\tikzstyle{txt2}=[draw,fill=white,inner sep=1pt,text centered,minimum width=1.0em, text height=1.5cm]

\newcommand{\includegraphicstmp}[3]{
\begin{tikzpicture}
%	\node[anchor=south west,inner sep=0, shift={(0,0)}] (image) at (0,0) {
%	\includegraphics[width=0.30\linewidth,trim=11.0cm 0.750cm 11.0cm 2.50cm,clip]{#1}
%	};
%	\begin{scope}[x={(image.south east)},y={(image.north west)}]
%		 \node [anchor=north east, draw, fill=none,thick, text width=2.5em,text centered, minimum height=2.50em] (box) at (0.70,0.880) {};	
%		 \node [anchor=north east, fill=white, text centered, above = -0.175cm of box]() {\footnotesize Car};	
%	 \end{scope}
	\node[anchor=south west,inner sep=0, shift={(0,0)}] (image) at (0,0) {
	\includegraphics[width=0.30\linewidth,trim=11.0cm 0.750cm 11.0cm 2.50cm,clip]{#1}
	};
	\begin{scope}[x={(image.south east)},y={(image.north west)}]
		 \node [anchor=north east, draw, fill=none,thick, text width=1.75em,text centered, minimum height=2.50em] (box) at (0.59,0.510) {};	
		 \node [anchor=north east, fill=white, text centered, above = -0.015cm of box]() {\footnotesize Cyclist};	
	 \end{scope}	 
	\node[anchor=south west,inner sep=0, shift={(0,3.0)}] (image2) at (0,0) {
	\includegraphics[width=0.30\linewidth,trim=11.0cm 0.750cm 11.0cm 3.50cm,clip]{#3}};
	\begin{scope}[x={(image2.south east)},y={(image2.north west)}]
	 \node [anchor=north east, draw, fill=none,thick, text width=1.25em,text centered, minimum height=2.450em] (box2) at (0.70,0.480) {};
		 \node [anchor=north east, fill=white, text centered, above = -0.005cm of box2]() {\footnotesize Ped.};	
 \end{scope}
\end{tikzpicture}
}%

%------------------------------------------ initial submission
%\newcommand{\pathtmpped}{./fig_predict_results_denoising/semseg_di_32_400_aug_both_sprinkles_full_LiLaBlockDil_1e3_2k5_ckp_40k_freeze/2018-11-26_153228_Dynamic-G2-Fog20_LidarImage_000000267} %
%
%\newcommand{\pathtmpcar}{./fig_predict_results_denoising/semseg_di_32_400_aug_both_sprinkles_full_LiLaBlockDil_1e3_2k5_ckp_40k_freeze/grey/2018-11-26_155631_Dynamic-G6-Fog30_LidarImage_000000284} %
%\newcommand{\pathtmpcyc}{./fig_predict_results_denoising/semseg_di_32_400_aug_both_sprinkles_full_LiLaBlockDil_1e3_2k5_ckp_40k_freeze/2018-11-26_153359_Dynamic-G4-Fog20_LidarImage_000000232} %
%
%%------------------cyclist
%\renewcommand{\pathtmpcyc}{./fig_predict_results_denoising/semseg_di_32_400_aug_both_sprinkles_full_LiLaBlockDil_1e3_2k5_ckp_40k_freeze/2018-11-26_153359_Dynamic-G4-Fog20_LidarImage_000000237} %

%------------------------------------------ revise and resubmit

%% file: img/fig_predict_results_denoising/denoise_results_dynamic.tex
\begin{tikzpicture}
 \node[anchor=south west,inner sep=0, shift={(0,0)}] (image) at (0,0) %
 {
 	%\subfloat[Raw Data]{\includegraphicstmp{%
	 %\pathtmpcyc_gt.png}{Input}{\pathtmpped_gt.png}}
 
	\subfloat[Raw Data]{\includegraphicstmp%
	{\weatherNetInitSub\pathCycA_gt.png}{Segmentation}%
	{\weatherNetInitSub\pathPedA_gt.png}}
	\subfloat[\textit{WeatherNet}]{\includegraphicstmp%
	{\weatherNetInitSub\pathCycA_prediction_solid.png}{Segmentation}%
	{\weatherNetInitSub\pathPedA_prediction_solid.png}}
	\subfloat[De-Noised]{\includegraphicstmp%
	{\weatherNetInitSub\pathCycA_denoised.png}{Segmentation}%
	{\weatherNetInitSub\pathPedA_denoised.png}}
};
\begin{scope}[x={(image.south east)},y={(image.north west)}]

% legend input
\node [box, fill=none, minimum height=14.5em] 	(box1) at (0.5,0.52) {};
\node [box, text width=7.5cm, above = 0.2cm of box1] 			(box2) {};

\node [txt, left = -1.5cm of box2] 								(txt1) {Input};
\node (l1box) [rectangle, left=0.05cm of txt1, minimum width=0.25cm, color=white, fill=colorInput, draw] {};
% segmentation
\node [txt, right = 0.6cm of txt1] 					(txt2) {Valid};
\node (l4box) [rectangle, left=0.05cm of txt2, minimum width=0.25cm, color=white, fill=colorValid, draw] {};

\node [txt, right = 0.6cm of txt2] 					(txt3) {Fog};
\node (l4box) [rectangle, left=0.05cm of txt3, minimum width=0.25cm, color=white, fill=colorFog, draw] {};

\node [txt, right = 0.6cm of txt3] 					(txt4) {Rain};
\node (l4box) [rectangle, left=0.05cm of txt4, minimum width=0.25cm, color=white, fill=colorRain, draw] {};

% de-noised
\node [txt, right = 0.6cm of txt4] 					(txt5) {De-Noised};
\node (l4box) [rectangle, left=0.05cm of txt5, minimum width=0.25cm, color=white, fill=colorDenoised, draw] {};

\end{scope}
\end{tikzpicture}

%% file: tex/related_work.tex
\info{Lidar in adverse weather, need of filter}
Adverse weather conditions such as fog, rain, dust or snow have a huge impact on the perception of lidar sensors, as
shown in \cite{Bijelic2018,Heinzler2019,Filgueira2017,Hasirlioglu2016,Hasirlioglu2017,Hasirlioglu2017a,Kutila2016,
Kutila2018,Lehtinen2018,Rasshofer2011,Phillips2017}. Consequently, point cloud processing algorithms either have to
deal with these influences, or require preprocessing by filter algorithms. Nevertheless, only a few de-noising algorithms 
for sparse point cloud data have been developed or are publicly available yet~(\cite{Shamsudin2016,Charron2018}). 
Most state-of-the-art data sets are recorded mostly under favorable weather conditions 
only~(e.g. \cite{AndreasGeiger2013,GauravPandey2011,Caesar2019}).%
\subsection{Dense Point Cloud De-Noising}
\info{{2D Image Filter and dense point cloud filter}}
Previous work on 2D depth image de-noising is mainly based on dense depth information obtained by stereo vision and
depth cameras (e.g. Intel RealSense, Microsoft Kinect, etc.) \cite{Carfagni2017,Shen2013}. Hence, traditional
algorithms developed over years for camera image de-noising can be applied in a straightforward fashion. These
approaches can be split in three different categories: (1)  spatial, (2) statistical and (3) segmentation-based
methods. %Image de-noising methods based on color information are not considered, as lidar point clouds are colorless.

\info{rel. work on spatial \cite{Ronnback2008, Li2014a}}%
Spatial smoothing filters (1), e.g. the Gaussian low pass filter, calculate a weighted average of pixel
values in the vicinity, where the weight decreases with the distance to the observed pixel. Points are smoothed by 
increasing distance from the derived weight \cite{Tomasi1998}.
For de-noising 2D point cloud data corrupted by snow, these filter types are providing successful results, as shown
by \cite{Ronnback2008} with a median filter. However by only assuming only small variations in the neighborhood, this approach
generally fails at preserving edges. The bilateral filter, introduced by \cite{Tomasi1998} for gray and color images,
is replacing traditional low-pass filtering by providing an edge preserving smoothing filter for dense depth
images~\cite{Li2014a}. \info{rel. work on statistical based: \cite{Jenke2006, Schall2005}}%

Statistical filter methods (2) for dense point cloud de-noising are often based on maximum likelihood
estimation~\cite{Schall2005} or Bayesian statistics \cite{Jenke2006}. By optimizing the decision whether a points lies
on a surface or not these approaches are smoothing surfaces and remove minor sensor errors.

\info{rel. work on segmentation based: \cite{Shen2013,Chen2012,Le2014}}%
By applying a segmentation step before filtering, segment-based filters (3) are smoothing only local segments of
point clouds with identical labels. Therefore corners and finer structures are better preserved. Region growing
\cite{Chen2012}, a maximum a-posteriori estimator \cite{Shen2013} or edge detection \cite{Le2014} is used for
segmentation, while bilateral filters are used for smoothing local segments.

\info{Why de-noising is different for lidar and depth image}
Lidar point clouds are significantly less dense compared to camera images, particularly at larger distances. 
As such, the direct application of camera algorithms does typically not achieve the desired result, 
as exemplified in \cite{Charron2018} for a median filter applied to point cloud data. 
Since conventional lidars have a resolution of a tenths of a degree and a range of two to three hundred meters, 
the density of the point cloud decreases significantly in the middle and far range.

A first machine learning approach for de-noising dense point clouds corrupted by fog with a visibility of $2\,$m and $6\,$m is introduced in \cite{Shamsudin2016}. By manually extracting features, a $k$ nearest neighbor (kNN) and a support vector machine (SVM)
are trained. The feature vector is in particular based on the eigenvalues of the covariance matrix of the Cartesian coordinates, therefore
it is only derived if there are more then ten points in a $50\,$mm$^{3}$ cubic voxel. 
For a sparse lidar point cloud this assumption is rarely satisfied.

\subsection{Sparse Point Cloud De-Noising}
\info{Lidar de-noising in 3D}
In the 3D domain many approaches are based on the spatial vicinity or statistical distributions of the point
cloud~\cite{Rusu2011}, such as the statistical outlier removal (SOR) and radius outlier removal (ROR) filter. The SOR
defines the vicinity of a point based on its mean distance to all $k$ neighbors compared with a threshold derived by
the global mean distance and standard deviation of all points. The ROR filter directly counts the number of neighbors
within the radius $r$ in order to decide whether a point is filtered or not.
Recently, Charron et al. \cite{Charron2018} have shown that these filter types are not suited for the de-noising task
of sparse point clouds corrupted by snow. Thus the enhanced dynamic radius outlier removal (DROR) filter was
introduced by \cite{Charron2018} which increases the search radius $r$ for neighboring points with increasing
distance of the measured point. Since this approach takes the raw data structure of lidar sensors into account, which
is less dense at far distances, a better performance could be achieved.

\info{Why Lidar de-noising with neighboring approaches failes}
Nevertheless, these approaches are based on spatial vicinity and consequently discard single reflections without
points in the neighborhood. As a result, points at greater distances are increasingly filtered, as shown
in~\cite{Charron2018} for the SOR, ROR and even DROR. Hence, valuable information for an autonomous vehicle,
especially at higher speeds, is discarded and the sensor's range is additionally limited by the filter.

\info{Neighboring approaches fail with dense water drops}
In addition, sparsity is not a valid feature to filter scatter caused by fog or drizzle, as soon as the density of
the distribution of water drops increases (Fig. \ref{fig:denoise_results_Dynamic-G2-Fog20}). In conclusion, we argue that these
filter approaches are prone to failure in the near and far range, as only spatial neighborhood is used.

\subsection{Semantic Segmentation for Sparse Point Clouds}
\info{Why Lidar de-noising with CNN, and explain CNN approaches}
We propose a filter approach based on a convolutional neural network, which understands the underlying data structure
and can generalize its characteristics for various distances and clutter distributions. Furthermore, this approach is
able to also incorporate the intensity information of the point cloud. The semantic segmentation task is being further
developed by a large scientific community and is already applied to the lidar point cloud domain, showing very
promising results \cite{Wu2017,Piewak2019b,Milioto2019}. A major advantage is that the algorithms can generalize very well and
thus recognize objects at different distances and orientations.

There are various established approaches for the input data layer and the network structure itself, which we utilize and 
adapt to the task of semantic weather segmentation~\cite{Chen2017,Yang2018,
Lang2018,Li2017,Piewak2019b,Milioto2019}. Since preprocessing algorithms have strong requirements on computation speed, we
focus on 2D input layer approaches, which commonly use a bird’s eye view (BEV) \cite{Chen2017, Yang2018, Lang2018} or
an image projection view \cite{Li2017, Piewak2019b, Milioto2019}.

The recently introduced \textit{PointPillars} by \cite{Lang2018} is based on a feature extraction network, generating a pseudo
image out of the point cloud which is used as input for a backbone \textit{CNN}. The approach excels on the KITTI's
object detection challenge \cite{AndreasGeiger2013} in terms of detection performance and inference time. However,
the approach has not yet been applied to point-wise semantic segmentation but is in wide-spread use for object
detection only. Thus, we propose a 2D approach inspired by the \textit{CNN} architecture of \textit{LiLaNet}
\cite{Piewak2019b}. % which is currently the best listed 2D approach on the recently released \textit{SemanticKITTI}
%benchmark \cite{Behley2019,AndreasGeiger2013}.

%% file: tex/method.tex
\label{sec:method}
\subsection{Lidar 2D Images}
State-of-the-art lidar sensors commonly provide raw data in spherical coordinates with the radius $r$, azimuth
angle~$\phi$ and elevation angle $\theta$, often combined with an estimated intensity or echo pulse width of the
backscattered light. The used rotating lidar sensors (Velodyne 'VLP32c') contain $32$ vertically stacked send/receive
modules, which are rotating to obtain the $360^\circ$ scan. Similarly to \cite{Piewak2019b} we merge one scan to a
cylindrical depth image as a 2D matrix $M = (m_{i,j}) \in \mathbb{R}^{(n \times m)}$ where each row $i$ represents
one of the $32$ vertically stacked send/receive modules and each column $j$ one of the $1800$ segments over the full
$360^\circ$ scan with the corresponding azimuth angle $\phi$ and timestamp $t$. As a consequence we obtain the
distance matrix $D \in \mathbb{R}^{(n \times m)}$ and intensity image $I \in \mathbb{R}^{(n \times m)}$.\\

\subsection{Autolabeling for Noise Caused by Rain or Fog}
\label{subsec:autolabeling}
In order to evaluate the quality of a trained classification approach, ground truth annotations are essential. For
sparse lidar point clouds, the manual annotation task is very challenging and even more difficult for semantic
weather segmentation, where the decision is whether a point is caused by a water droplet or not. Human comprehension
of camera images is much more powerful than of lidar point clouds, therefore a time-synchronized camera image as
additional information is helpful for labeling lidar point clouds in order to significantly improve the label quality. 
However, since water droplets cannot be captured directly by passive camera sensors, especially at large distances, this
label aid is not available for semantic labeling of weather information. Thus, we utilize the recorded static scenes
in controlled environments to develop an automated labeling procedure, which does not involve human perception.

We stack all $f$ lidar images for each frame $k$ from one sensor in reference conditions to obtain one single point cloud
$D^{GT} = (d^{GT}_{i,j,k}) \in \mathbb{R}^{(n\times m \times f)}$. Subsequently, we compare each distance image $D$
captured during rain or fog with all ground truth images $D^{GT}$ to decide whether a point is labeled as clutter or
not. Since the reference measurements are accumulated over several frames, minor measurement inaccuracies of the
sensor are already taken into account when comparing the distance images.
In addition, a threshold $\Delta R$ is added to the search region of valid distances. The threshold value
$\Delta R=\pm35\,$cm was chosen rather high, compared to the specified distance precision of the sensor,
in order to minimize the number of false negatives. Hence, the labels are derived mathematically for each distance
$d_{i,j}$ with the corresponding ground truth vector for this element $d^{GT}_{i,j,k}$:
\begin{numcases}{p=}
	\mathrm{clutter}, & if $\Delta R \geq \smash{\displaystyle\min_{1 \leq k \leq f}} |d^{GT}_{i,j,k} - d_{i,j}|  $\\
	\mathrm{no \,\, clutter}, & else
\end{numcases}
By directly comparing the distances of the same transmitter and receiver pairs, this method is very fast, directly
based on the sensor raw data and does not require 3D information. Alternatively, a 3D point cloud
comparison was implemented by a kd-tree approach without showing significantly different results. In order to be able
to distinguish between different weather conditions, fog and rain sequences are provided with different labels. 
%The ground truth in Fig. \ref{fig:point_cloud_results} illustrates the results of the automated labeling approach.
\FPeval{\strongestFalsePointsMean}{clip(27.657092290727846)}%
\FPeval{\strongestFalsePointsStd}{clip(14.752476735418258)}%
\FPeval{\lastFalsePointsMean}{clip(46.944637950046385)}%
\FPeval{\lastFalsePointsStd}{clip(6.820618378962131)}%
\FPeval{\strongestFalsePointsPerPixelMean}{clip(27.657092290727846/(32*400)*100)}%
\FPeval{\strongestFalsePointsPerPixelStd}{clip(14.752476735418258/(32*400)*100)}%
\FPeval{\lastFalsePointsPerPixelMean}{clip(46.944637950046385/(32*400)*100)}%
\FPeval{\lastFalsePointsPerPixelStd}{clip(6.820618378962131/(32*400)*100)}%
\FPeval\resultSTDgt{round(\lastFalsePointsPerPixelStd:3)}%
\FPeval\resultMeangt{round(\lastFalsePointsPerPixelMean:3)}%

To quantify the error of our ground truth labeling, we applied the label procedure on the reference recording itself.
We split the recording during reference for one setup into equally sized parts. The evaluation is done by
taking the accumulation of the first split as valid points for labeling the second split and vise versa.
As there are no changes in weather conditions, we would expect that all points will be labeled as valid.
The evaluation demonstrates a mean per pixel false rate of $\resultMeangt\pm\resultSTDgt\,\%$ for both tests.%
\subsection{Data Augmentation}
%\begin{figure}[tb]
%	\centering
%	\begin{centering}
%		%	\resizebox{0.9\linewidth}{!}{\input{img/mean_intensity_fog_rain_points/mean_intensity_fog_points}}
%		\input{img/mean_intensity_fog_rain_points/mean_intensity_fog_points}
%		\caption{\DRAFT{Intensity distribution of points whose existence can be explained by weather influences  (e.g
%rain or fog). The data is based on the climate chamber recordings at $5-100\,m$ visbility and a rainfall rate of 
%$15$, $33$ and $55\,mm/h$. A logarithmic normal distribution $LN(\mu,\sigma^2)$ is assumed as underlying probability 
%density distribution function.}}
%		\label{fig:mean_intensity_weather_points}
%		%\mathcal{LN}
%	\end{centering}
%\end{figure}
%\begin{figure*}[tb]
%	\centering
%	\resizebox{0.85\linewidth}{!}{\input{img/cnn_network/encoder_decoder_block}}	
%	\caption{The \textit{Encoder} and \textit{DecoderBlock}. The Encoder Block is inspired by the RangeNet++
%\cite{Milioto2019} with the Leaky Relu activation function.}
%	\label{fig:cnn_encoder_decoder_block}
%\end{figure*}
\begin{figure}[tb]
	\centering
	\vspace{0.01cm}
	\resizebox{0.95\linewidth}{!}{\input{img/cnn_network/lila_block}}	
	\caption{The modified \textit{LiLaBlock} is based on \cite{Piewak2019b} and enlarged by a dilated convolution 
	\cite{Schall2005}.}\vspace{-0.2cm}
	\label{fig:lilablock}
\end{figure}
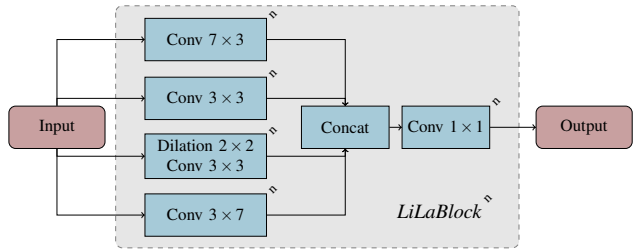
\begin{figure*}[tb]
	\centering
	\resizebox{1.0\linewidth}{!}{\input{img/cnn_network/weather_net}}
	\caption{The proposed \textit{WeatherNet} architecture is based on \textit{LiLaNet} introduced by 
	\cite{Piewak2019b} and optimized for the de-noising purpose. Therefore, the depth is reduced, a dropout layer is 
	inserted and a dilated convolution is added to base block of the network. The modified \textit{LiLaBlock} is 
	given in detail in Fig. \ref{fig:lilablock}.}\vspace{-0.2cm}
	\label{fig:cnn_weather_net}
\end{figure*}
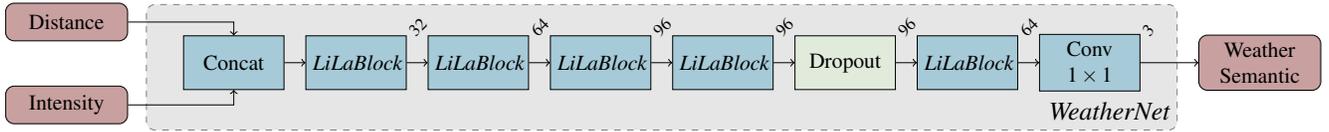
State-of-the-art sparse point cloud data sets which are publicly accessible tend to be recorded under favorable 
weather conditions. To be able to utilize these data sets for semantic weather segmentation, we developed an 
augmentation approach for rain based on the fog model of \cite{Bijelic2019}. Hence, we obtained a large training data set without requiring manual annotation while providing error-free ground truth. The augmentation algorithm is applied to lidar images to enable manipulations for each individual distance measurement, whereby occlusion is conceptually 
impossible. 
The proposed augmentation based on the model of \cite{Bijelic2019} does not only add individual points but alters additional attributes of the point cloud: Adverse weather affects viewing range and lowers the contrast of intensity and echo pulse widths, respectively.
%Lidar sensors are severely affected by adverse weather, which results in a lower viewing range, lower contrast in terms of intensity or echo pulse width and additionally water drops from fog or rain can be perceived and are present in the point cloud. The proposed fog model of \cite{Bijelic2019} aims to augment all of these disturbances. 
\subsubsection{Fog Model}
First of all the maximum sensing range is derived by the runway visual range $V = -ln(C_T)/\beta$ based on the atmospheric extinction coefficient $\beta$ and the observer’s contrast threshold $C_T$.
For lidar sensors $C_T$ can be interpreted as a detector threshold, where the 
sensor is able to perceive an object above the noise floor \cite{Bijelic2019}. As lidar is an active sensor system, 
the maximum sensor range is the half of the maximum viewing distance and results in:
\begin{equation}
	d_{max} = \frac{-ln(\frac{n}{L_{fog}+g})}{2\cdot\beta}
	\label{eq:runway_vis_range}
\end{equation}
The sensor threshold is a function of the received laser intensity $L_{fog}$, the adaptive laser gain $g$ and the 
detectable noise floor $n$. Scattering points due to waterdrops are added according to the model of 
\cite{Bijelic2019}. 
In contrast to \cite{Bijelic2019} the intensities of augmented points are derived from a logarithmic normal distribution $LN(\mu,\sigma^2)$, which is assumed as the underlying probability distribution function. The parameters $\mu$ and $\sigma$ are derived from the intensities of all clutter points based on the autolabeled climate chamber data from the
previous section. Hence we are able to model the intensity distribution of fog between $10-100\,m$ meteorological visibility and for
rainfall rates at $15$, $33$ and $55\,mm/h$. We preferred this method because in the model of \cite{Bijelic2019} the 
original scene is mirrored in the intensity distribution, as the augmented intensities $\tilde{I}$ are a function of
the perceived intensities $I$ of the sensor ($\tilde{I}=I\cdot e^{-\beta \cdot d}$).
The augmented fog corresponds to a visibility $V$ of $30-3000\,\mathrm{m}$, hence we use a atmospheric extinction coefficient $\beta$ 
between $0.001$ and $0.1$.\\
\setlength{\textfloatsep}{1pt}
\begin{algorithm}[tb]
\begin{footnotesize}%
	%	\KwData{this text}
	%	\KwResult{Result}
	\SetKwFunction{KwFn}{Fn} 
	\SetKwFunction{FMain}{Main}
	\SetKwProg{Fn}{Function}{}{end}
	\Fn{rain($D, I, \beta, p$)}{
		$B = $betafunction$(\beta)$\\
		$D_{max}=-ln(\frac{n}{I+g})/(2B)$\\
%		$D_{scatter}=-ln(0.5)/(B)$\\
		$D_{rand}=$random.uniform$(D_{max}$)\\
		$P_{lost}=1-exp(-\beta \cdot D_{max})$\\
		\ForEach{$d\in D, d_{m}\in D_{max}, d_{s}\in D_{scatter}, d_{r}\in D_{rand}, p_{l}\in P_{lost}$}{
			\eIf{$d_{m}<d$}
			{	
				\uIf{$p_{l}$}{
					pass\Comment*[r]{point is lost, do nothing}
				}
%				\uElseIf{$d_{s}<d$}{
%					$d = d_{s}$\Comment*[r]{scatter point}
%					$i = LN(\mu,\sigma^2)$\Comment*[r]{rain int.}
%				}
				\uElseIf{rand$<p$}
				{
					$d = d_{r}$\Comment*[r]{random scatter point}
					$i = LN(\mu,\sigma^2)$\Comment*[r]{rain int.}
				}
				\uElse
				{
						pass
				}  
				
			}
			{
				$i = i \cdot exp(-\beta \cdot d)$\Comment*[r]{attenuate int.}
			}
		}
		\KwRet $D$, $I$
	}
	\label{alg:rain_augmentation}
	\caption{{\footnotesize Point cloud rain augmentation model $rain(D, I, \beta, R)$ with distance matrix $D$, intensity matrix $I$, atmospheric 
	extinction coefficient $\beta$ and point scatter rate $p$.}} %\vspace{-0.5cm}%
\end{footnotesize}%
\end{algorithm}%
\subsubsection{Rain Model}
Besides our modifications of the fog augmentation based on \cite{Bijelic2019}, we further developed a 
rain augmentation. Thereby the parameters from \cite{Bijelic2019} are adapted to make the augmented scatter points equivalent to natural rainfall. The atmospheric extinction coefficient $\beta$ is set to $0.01$ for rain augmentation. The point scatter rate $p$ defines the per point probability of random scatter points.
The obtained point-wise ground truth data enables the calculation of $p$ for raindrops, which is $10.61\,\%$, $0.73\,\%$ and $4.70\,\%$ for $15$, $33$ and $55\,$ mm/h in the climate chamber. For the augmentation we finally fixed $p$ at $7.5\,\%$, which stabilizes the \textit{CNN} training, matches the quantity of scatter points in natural rainfall and is in the range of the derived probabilities from the climate chamber. The rain augmentation is described in detail in Algorithm \ref{alg:rain_augmentation}.
\subsection{Network Architecture}
\info{idea: semantic weather segmentation}
For the de-noising of lidar images we adopt state-of-the-art \textit{CNN} architectures for semantic segmentation 
of sparse point clouds. The proposed \textit{WeatherNet} is an efficient variant of the \textit{LiLaNet} introduced 
by \cite{Piewak2019b}. In order to optimize the network for the de-noising task, we reduced the depth of the
network given that the complexity of our task ($3$ classes) is reduced in comparison to full multi-class semantic 
segmentation ($13$ classes) \cite{Cordts2016, Piewak2019b}. 
%As a result, inference of our \textit{WeatherNet} on VLP32C point clouds is done in \TODO{<1~ms} on an \TODO{NVIDIA GTX XXXX GPU}. 

Additionally, we adapted the inception layer (Fig. \ref{fig:lilablock}) to include a dilated convolution to provide more  
information about the spatial vicinity by increasing the receptive field. Further, a dropout layer is inserted to 
increase the capability of generalization. The resulting network architecture is illustrated in Fig. 
\ref{fig:cnn_weather_net}.
After optimization on the validation data set, the batch size is set to $b=20$, the learning rate to 
$\alpha=4\cdot10^{-8}$ with a learning rate decay of $0.90$ after every epoch. Adam solver is used to perform the 
training, with the suggested default values $\beta_1=0.9$, $\beta_2=0.999$ and $\epsilon=10^{-8}$ \cite{Kingma2014}.%

%% file: img/cnn_network/lila_block.tex
% -*- root: ../../root.tex -*-
%
% Define the layers to draw the diagram
\pgfdeclarelayer{network}
\pgfsetlayers{network,main}

\tikzstyle{nin}=[draw, fill=daiPetrol7, text width=6.5em,text centered, minimum height=2.50em]
\tikzstyle{wa} = [nin, text width=5.0em, fill=daiDeepRed7,minimum height=2.5em, rounded corners]

% Define distances for bordering
\def\blockdist{1.3}
\def\edgedist{1.0}

\begin{tikzpicture}[trim left]

%------------------------------------------------------------------------
% Input
%------------------------------------------------------------------------

\node (input) [wa] {Input};

%------------------------------------------------------------------------
% Convs
%------------------------------------------------------------------------

\node (horizontal) at ([shift={(1.6*\blockdist,1.4*\blockdist)}]input.east)[nin]  {Conv $7\times3$};
\path [draw, ->] (input.north) |- node [above] {} (horizontal.west);
\node[label={[rotate=45]right:\footnotesize n}] at ([shift={(-0.1*\blockdist,0.3*\blockdist)}]horizontal.east) {};

\node (square) at ([shift={(1.6*\blockdist,0.6)}]input.east)[nin]  {Conv $3\times3$};
%\path [draw, ->] (input.east) -- node [above] {} (square.west);
\path [draw, ->] (input.north) |- node [above] {} (square.west);
\node[label={[rotate=45]right:\footnotesize n}] at ([shift={(-0.1*\blockdist,0.3*\blockdist)}]square.east) {};

\node (dilated) at ([shift={(1.6*\blockdist,-0.6)}]input.east)[nin]  {Dilation $2\times2$ Conv $3\times3$};
%\path [draw, ->] (input.east) -- node [above] {} (dilated.west);
\path [draw, ->] (input.south) |- node [above] {} (dilated.west);
\node[label={[rotate=45]right:\footnotesize n}] at ([shift={(-0.1*\blockdist,0.3*\blockdist)}]dilated.east) {};

\node (vertical) at ([shift={(1.6*\blockdist,-1.4*\blockdist)}]input.east)[nin]  {Conv $3\times7$};
\path [draw, ->] (input.south) |- node [above] {} (vertical.west);
\node[label={[rotate=45]right:\footnotesize n}] at ([shift={(-0.1*\blockdist,0.3*\blockdist)}]vertical.east) {};

%------------------------------------------------------------------------
% Concat
%------------------------------------------------------------------------

\node (concat) at ([shift={(1.25*\blockdist,-0.6)}]square.east) [nin, text width=4.5em, minimum height=2.5em]{Concat};
\path [draw, ->] (horizontal) -| node [above] {} (concat.north);
%\path [draw, ->] (square) -- node [above] {} (concat.west);
\path [draw, ->] (square) -| node [above] {} (concat.north);
\path [draw, ->] (dilated) -| node [above] {} (concat.south);
\path [draw, ->] (vertical) -| node [above] {} (concat.south);

%------------------------------------------------------------------------
% Bottleneck
%------------------------------------------------------------------------

\node (bottleneck) at ([shift={(0.9*\blockdist,0.0)}]concat.east)[nin, text width=4.5em, minimum height=2.5em]  {Conv $1\times1$};
\path [draw, ->] (concat.east) -- node [above] {} (bottleneck.west);
\node[label={[rotate=45]right:\footnotesize n}] at ([shift={(-0.1*\blockdist,0.3*\blockdist)}]bottleneck.east) {};

%------------------------------------------------------------------------
% output
%------------------------------------------------------------------------

\node (output) at ([shift={(1.5*\blockdist,0)}]bottleneck.east)[wa] {Output};
\path [draw, ->] (bottleneck.east) -- node [above] {} (output.west);

%------------------------------------------------------------------------
% Layout
%------------------------------------------------------------------------

\begin{pgfonlayer}{network}
\path[fill=daiCoolGrey,rounded corners, draw=black!50, dashed]
(horizontal.west)+(-0.6*\edgedist,0.70*\edgedist) rectangle ([shift={(0.6*\edgedist,-2.5*\edgedist)}]bottleneck.east);
\end{pgfonlayer}

\path (bottleneck.south) +(-0.1*\blockdist,-1.0*\blockdist) node (lilablock) {\large{\emph{LiLaBlock}}};
\node[label={[rotate=45]right:\footnotesize n}] at ([shift={(-0.2*\blockdist,0.1*\blockdist)}]lilablock.east) {};

\end{tikzpicture}

%% file: img/cnn_network/weather_net.tex
% -*- root: ../../root.tex -*-
%
% Define the layers to draw the diagram
\pgfdeclarelayer{network}
\pgfsetlayers{network,main}

\tikzstyle{nin}=[draw, fill=daiPetrol7, text width=4.0em,text centered, minimum height=2.50em]
\tikzstyle{drop}=[draw, fill=daiGreen4, text width=4.0em,text centered, minimum height=2.50em]
\tikzstyle{decoder}=[nin, fill=daiGreen4, trapezium, rotate=-90, trapezium left angle=70, trapezium right angle=70,  minimum height=2.0em, minimum width=2.5em, text width=1.0em]
\tikzstyle{encoder}=[decoder, rotate=180]
\tikzstyle{ende}=[draw, fill=daiGreen4, text width=4.0em,text centered, minimum height=2.5em]
\tikzstyle{wa} = [nin, text width=5.0em, fill=daiDeepRed7,minimum height=1.75em, rounded corners]

%draw,trapezium,trapezium left angle=70,trapezium right angle=0

% Define distances for bordering
\def\blockdist{1.3}
\def\edgedist{1.0}

\begin{tikzpicture}[trim left]

%------------------------------------------------------------------------
% Input
%------------------------------------------------------------------------

\node (distance_input) [wa] {Distance};
%\path (distance_input.south) +(0.0*\blockdist,-0.3*\blockdist) node (ditance_image) {\includegraphics[width=2cm, height = 5mm]{img/cnn_network/lidar_dist_img_example_fog53.png}};
\node (intensity_input) at ([shift={(0*\blockdist,-0.80*\blockdist)}]distance_input.south) [wa] {Intensity};
%\path (intensity_input.north) +(0.0*\blockdist,0.3*\blockdist) node (intensity_image) {\includegraphics[width=2cm, height = 5mm]{img/cnn_network/lidar_int_img_example_fog53.png}};

%------------------------------------------------------------------------
% Concat
%------------------------------------------------------------------------

\node (concat) at ([shift={(2.1*\blockdist,-0.35)}]distance_input.south) [nin]  {Concat};
\path [draw, ->] (distance_input) -| node [above] {} (concat.north);
\path [draw, ->] (intensity_input) -| node [above] {} (concat.south);

%------------------------------------------------------------------------
% LiLaBlocks
%------------------------------------------------------------------------

\node (llb1) at ([shift={(0.9*\blockdist,0)}]concat.east)[nin]  {\emph{LiLaBlock}};
\path [draw, ->] (concat.east) -- node [above] {} (llb1.west);
\node[label={[rotate=45]right:\footnotesize32}] at ([shift={(-0.1*\blockdist,0.3*\blockdist)}]llb1.east) {};

%\node (decoder1) at ([shift={(0.6*\blockdist,0)}]llb1.east)[decoder] {\rotatebox{90}{\textit{\textit{En}}}};
%\path [draw, ->] (llb1.east) -- node [above] {} (decoder1.south);
%\node[label={[rotate=45]right:\footnotesize64}] at ([shift={(0.1*\blockdist,0.6*\blockdist)}]decoder1.east) {};

\node (llb2) at ([shift={(0.9*\blockdist,0)}]llb1.east)[nin]  {\emph{LiLaBlock}};
\path [draw, ->] (llb1.east) -- node [above] {} (llb2.west);
\node[label={[rotate=45]right:\footnotesize64}] at ([shift={(-0.1*\blockdist,0.3*\blockdist)}]llb2.east) {};

%\node (decoder2) at ([shift={(0.6*\blockdist,0)}]llb2.east)[decoder] {\rotatebox{90}{\textit{En}}};
%\path [draw, ->] (llb2.east) -- node [above] {} (decoder2.south);
%\node[label={[rotate=45]right:\footnotesize64}] at ([shift={(0.1*\blockdist,0.6*\blockdist)}]decoder2.east) {};

\node (llb3) at ([shift={(0.9*\blockdist,0)}]llb2.east)[nin]  {\emph{LiLaBlock}};
\path [draw, ->] (llb2.east) -- node [above] {} (llb3.west);
\node[label={[rotate=45]right:\footnotesize96}] at ([shift={(-0.1*\blockdist,0.3*\blockdist)}]llb3.east) {};

\node (llb4) at ([shift={(0.9*\blockdist,0)}]llb3.east)[nin]  {\emph{LiLaBlock}};
\path [draw, ->] (llb3.east) -- node [above] {} (llb4.west);
\node[label={[rotate=45]right:\footnotesize96}] at ([shift={(-0.1*\blockdist,0.3*\blockdist)}]llb4.east) {};

\node (dropout) at ([shift={(0.9*\blockdist,0)}]llb4.east)[drop]  {Dropout};
\path [draw, ->] (llb4.east) -- node [above] {} (dropout.west);
\node[label={[rotate=45]right:\footnotesize96}] at ([shift={(-0.1*\blockdist,0.3*\blockdist)}]dropout.east) {};

\node (llb5) at ([shift={(0.9*\blockdist,0)}]dropout.east)[nin]  {\emph{LiLaBlock}};
\path [draw, ->] (dropout.east) -- node [above] {} (llb5.west);
\node[label={[rotate=45]right:\footnotesize64}] at ([shift={(-0.1*\blockdist,0.3*\blockdist)}]llb5.east) {};

%\node (encoder2) at ([shift={(0.6*\blockdist,0)}]llb5.east)[encoder] {\rotatebox{-90}{\textit{De}}};
%\path [draw, ->] (llb5.east) -- node [above] {} (encoder2.north);
%\node[label={[rotate=45]right:\footnotesize64}] at ([shift={(0.1*\blockdist,-0.3*\blockdist)}]encoder2.east) {};

%\node (encoder1) at ([shift={(0.6*\blockdist,0)}]encoder2.south)[encoder] {\rotatebox{-90}{\textit{De}}};
%\path [draw, ->] (encoder2.south) -- node [above] {} (encoder1.north);
%\node[label={[rotate=45]right:\footnotesize64}] at ([shift={(0.1*\blockdist,-0.3*\blockdist)}]encoder1.east) {};

\node (final) at ([shift={(0.9*\blockdist,0)}]llb5.east)[nin]  {Conv $1 \times 1$};
\path [draw, ->] (llb5.east) -- node [above] {} (final.west);
\node[label={[rotate=45]right:\footnotesize3}] at ([shift={(-0.1*\blockdist,0.3*\blockdist)}]final.east) {};

%------------------------------------------------------------------------
% skip
%------------------------------------------------------------------------
%\draw [draw, ->] (decoder1.west)    -- +(0,.75)  -| (encoder1.east);
%\draw [draw, ->] (decoder2.west)    -- +(0,.45)  -| (encoder2.east);
%------------------------------------------------------------------------
% output
%------------------------------------------------------------------------
\node (output) at ([shift={(1.5*\blockdist,0)}]final.east)[wa] {Weather Semantic};
\path [draw, ->] (final.east) -- node [above] {} (output.west);

%\path (output.north) +(0.0*\blockdist,0.3*\blockdist) node (output_image) {\includegraphics[width=2cm, height = 5mm]{img/cnn_network/lidar_label_img_example_fog53.png}};

%------------------------------------------------------------------------
% Layout
%------------------------------------------------------------------------
\begin{pgfonlayer}{network}
\path[fill=daiCoolGrey,rounded corners, draw=black!50, dashed]
(concat.west)+(-0.6*\edgedist,0.95*\edgedist) rectangle ([shift={(0.6*\edgedist,-1.10*\edgedist)}]final.east);
\end{pgfonlayer}

\path (final.south) +(0.25*\blockdist,-0.25*\blockdist) node (lilanet) {\large{\emph{WeatherNet}}};

\end{tikzpicture}

%% file: tex/dataset.tex
\label{sec:dataset}
\setlength{\textfloatsep}{20pt}
\newcommand{\imgchamberpath}[1]{\includegraphics[trim=3cm 11.5cm 2.0cm 1.80cm,clip,width=0.333\linewidth]%
	{img_static_scenes_climate_chamber/#1}}
\begin{figure}[tb]
	%trim=left bottom right top, clip]
%	\vspace{0.01cm}
%	\subfloat[][Pedestrian Crossing]{\includegraphics[width=1.0\linewidth]%
%	{img_static_scenes_climate_chamber/static_scene_2.pdf}
%		\label{fig:IMG_2205_Static2_pedestrian_crossing}}\\[-2ex]		
%	\subfloat[][Construction Area]{\includegraphics[trim=7.5cm 11.0cm 12.0cm 7.0cm,clip,width=0.333\linewidth]%
%	{img_static_scenes_climate_chamber/IMG_2160_Static3_construction_site_print}%
%		\label{fig:IMG_2160_Static3_construction_site}}
%	\subfloat[][Highway]{\includegraphics[trim=2cm 5.0cm 4.5cm 5.0cm,clip,width=0.333\linewidth]%
%	{img_static_scenes_climate_chamber/IMG_2158_Static4_Highway_flip_print}%
%		\label{fig:IMG_2158_Static4_Highway}}		
%	\subfloat[][Pedestrian Zone]{\includegraphics[trim=0.0cm 0.0cm 0.0cm 6.0cm,clip,width=0.333\linewidth]%
%	{img_static_scenes_climate_chamber/IMG_2223_Static1_pedestrian_zone_flip_print}%
%		\label{fig:IMG_2223_Static1_pedestrian_zone}}	
\includegraphics[width=1.0\linewidth]{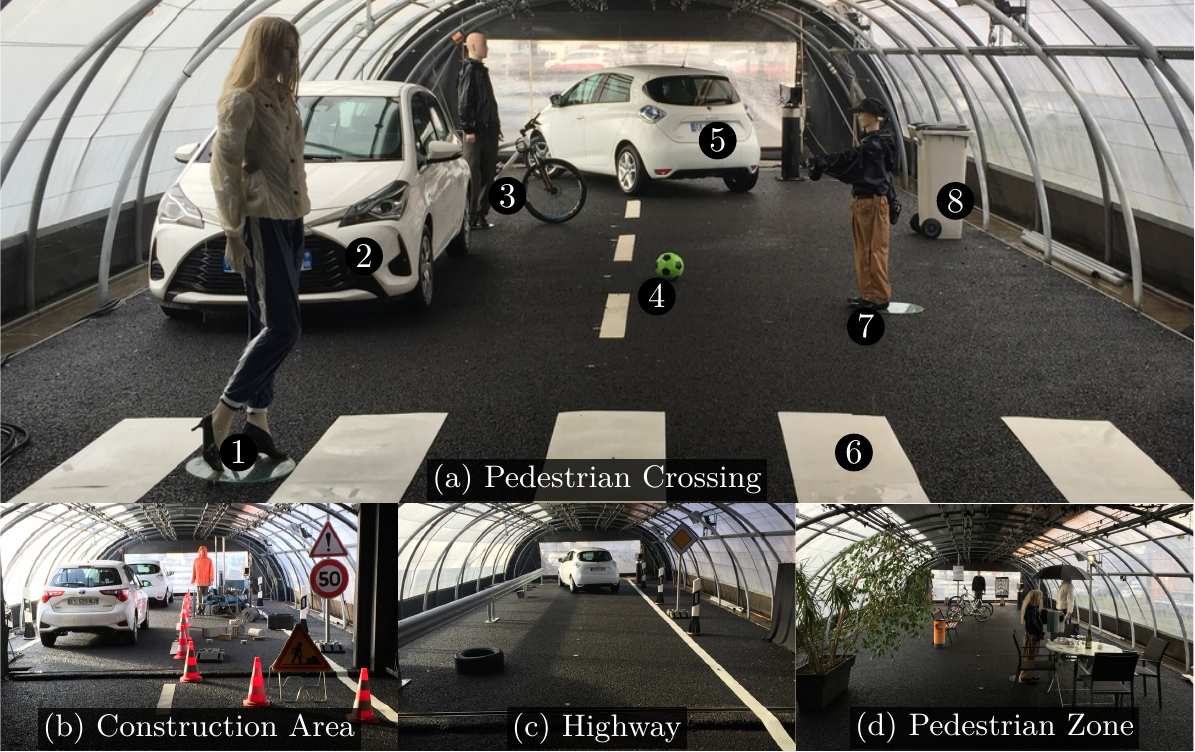}
	\caption{Static setups in the chamber representing four complex real traffic 
	situations. The upper picture shows a scene with a pedestrian (1) on a zebra crossing (6), a child (7) imitating chasing a ball (4) on the street, a parking car (2), a cyclist who pushes his bike across the street and a car (5) that turns left. 
	In addition there is a garbage can (8) rightmost. 
	The bottom pictures show several traffic scenarios 
%	a pedestrian zone area, a highway scenario and a construction area
	with various different objects like a black tire as lost cargo, guardrails, cars, lane 
	markings, reflector posts, traffic signs, a plant and pedestrian mannequins with and without umbrella.
	}\vspace{-0.3cm}
	\label{fig:chamber_static_scenes}
\end{figure}
\subsection{Road Data Set}
Creating a large-scale data set for training, validation and testing for the purpose of weather segmentation is very 
challenging, due to the fact that weather conditions are very unique and manual annotations are very difficult and 
complex. In order to re-use data sets, which were recorded under favorable weather conditions, like the data set from \cite{Piewak2019b}, we apply the developed data augmentation. Hence, we are able to utilize data sets recorded under favorable weather conditions with various traffic scenarios and roads types for the training of semantic weather segmentation. 
\begin{figure*}
	\centering
	\vspace{0.01cm}
	\includegraphics[width=0.47\linewidth]{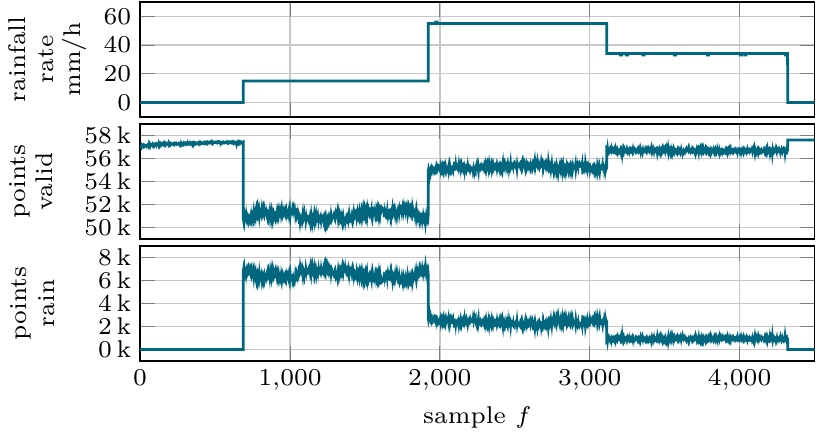}\hspace{0.25cm}
	\includegraphics[width=0.47\linewidth]{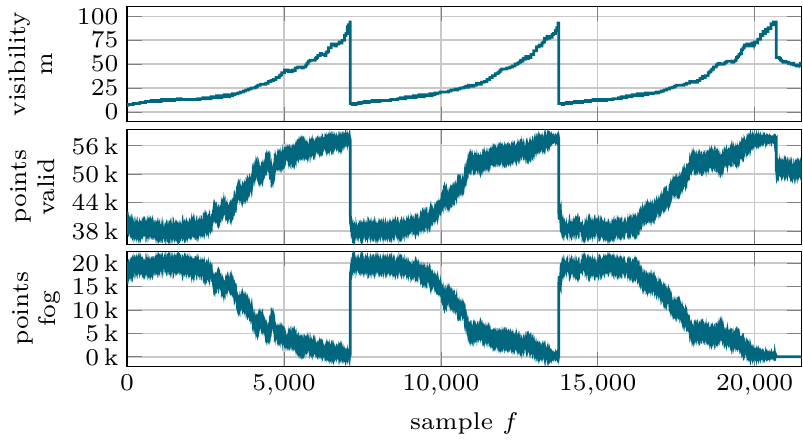}
	\caption{Illustration of the meteorological visibility in $m$ and rainfall rate in $mm/h$ provided by the climate
	chamber, the number of valid points and the number of scatter points during rainfall or fog.}\vspace{-0.10cm}
	\label{fig:fog_clear_pointwise}
\end{figure*}
\subsection{Climate Chamber Scenarios and Ground Truth Labels}
Furthermore, a large data set with four very realistic road scenarios was recorded in CEREMA's climatic chamber to 
obtain constant and reproducible fog, rain and reference conditions \cite{Gruber2019,Colomb2008a}. The data set will be published and will be available under the link in the abstract.
Fig. \ref{fig:chamber_static_scenes} illustrates these static scenes with various objects, 
which are intended to provide a remarkably realistic representation.
The climate chamber data set contains fog visibility ranges of 
$10-100\,\mathrm{m}$ and rain intensities of $15$, $33$ and $55\,\mathrm{mm/h}$. 
The rainfall rate is close-loop controlled at a constant level. 
The fog recording was started at a visibility of $10\,\mathrm{m}$ and recorded until
$100\,\mathrm{m}$ during a continuous dissipation and measurement of the actual visibility provided by a reference system of the climate chamber \cite{Colomb2008a}. 
This procedure enables very accurate determination of the visibility and was repeated three times to generate more 
samples for each meteorological visibility.

As described in Section \ref{sec:method}, the reference data recorded without any rainfall or fog, enables the 
proposed \textit{autolabeling} procedure. 
Therefore we obtained a large point-wise annotated data set without the error susceptibility of manually labeled 
weather data. 
The label set contains the classes 'clear' representing any point in the point cloud which is not caused by 
adverse weather, 'rain' for rain clutter and 'fog' for fog clutter respectively. %

Based on the data set, we are able to analyze the weather influence on a point level, as illustrated in Fig.~\ref{fig:fog_clear_pointwise}.
The analysis shows that the lidar point cloud reflects the weather conditions in a very detailed level, since the
number of points scattered by fog or rain is correlated with the visibility or rainfall rate.
In addition, the results indicate that no points are being lost and therefore the sum of fog or rain and valid points is
equivalent to the number of points in reference conditions.
Thus the point cloud contains the information to estimate the meteorological visibility or rainfall rate by determining the number of weather induced scattering points. As an increase in the rainfall rate does not necessarily
results in an increase of scatter points, the rainfall rate cannot be estimated directly, but the extent of the
degradation of the lidar sensor can be estimated.
This information is incredibly valuable for an autonomous vehicle to adapt behavior to environmental conditions and 
sensor performance. %
%%%%%%%%%%%%%%%%%%%%%%%%%%%%%%%%%%%%%%%%%%%%%%%%%%%%%%%%%%%%%%%%%%%%%%%%%%%%%%%%%%%%%%%%%%%%%%%%%%%%%%%%%%%%%%%%%%%%%%%%%%%%%%%%
\subsection{Data Split}
\FPeval{\chamberSamples}{clip(72800)}%
\FPeval{\roadSubset}{clip(31078)}%
%\FPeval{\roadSamples}{clip(555035)}%overall (exp 5)
\FPeval{\roadSamples}{clip(103141)}% exp03
\FPeval{\overallSamples}{clip(\roadSamples+\chamberSamples)}%
\FPeval{\overallTrainSamples}{clip(343926+7004+13553+13923)}%
\FPeval{\chamberRoadSubset}{clip(\chamberSamples+\roadSubset)}%
In total, the data set contains about %
$\num{\overallSamples}$ samples for training, validation and testing containing chamber ($\num{\chamberSamples}$) and road ($\num{\roadSamples}$) scenes, which 
can be used thanks to augmentation. Details about the number of samples and class distributions are stated in Table \ref{table:overall_results}. 
%An overview of the utilized data set and the number of frames for different classes and recorded scenarios is
%stated in Fig. \ref{fig:dataset_statistics}. 
In order to reduce time correlations between samples which were
recorded in the climate chamber, each setup is only used in the training (Fig.
\ref{fig:chamber_static_scenes}d, \ref{fig:chamber_static_scenes}c), validation (Fig.
\ref{fig:chamber_static_scenes}b) or test data split (Fig. \ref{fig:chamber_static_scenes}a). In total we obtain a data split of about ($60\%-15\%-25\%$) for training, validation and test.

To reduce an over-fitting to local dependencies and the scenes in the climate chamber, we cropped the image for
training to a forward facing view of about $60^\circ$ in the horizontal field of view.
In addition, a subset of the data set, already used in \cite{Piewak2019b}, is added as samples in favorable weather conditions to increase the diversity and add road recordings while maintaining a balanced class distribution.
Thus, the \textit{'chamber $\&$ road'} data set contains $\num[group-separator={,}]{\chamberRoadSubset}$ and $\num[group-separator={,}]{\roadSubset}$ road samples without augmentation or adverse weather.
Further, the amount and diversity of the training data can be increased by a manifold, because the augmentation enables the utilization of large-scale road data set, which results in $\num[group-separator={,}]{\roadSamples}$ road samples and in $\num{\overallSamples}$ in total.% and $\num[group-separator={,}]{\overallSamples}$ in total.
%\setlength{\tabcolsep}{3pt}
%\begin{table}[tb]
%	\centering
%	\caption{Overview about all experients and the utilized training, validation and test data set split}
%	\label{table:training_experiments}
%	\begin{tabular}{lp{3.25cm}P{0.75cm}P{0.75cm}P{0.75cm}}
%		\toprule
%		$\#$ & setup name 		&  train & val 	& test  \\
%		\midrule
%		D& pedestrian zone 		& 	$\bullet$ 	& $\circ$ 	& $\circ$\\
%		C& highway 				& 	$\bullet$ 	& $\circ$ 	& $\circ$\\\midrule
%		B& construction area 		& 	$\circ$ 	& $\bullet$ & $\circ$\\\midrule
%		A& pedestrian crossing 	& 	$\circ$ 	& $\circ$ 	& $\bullet$\\\bottomrule
%	\end{tabular}
%\end{table}%

%% file: tex/experiments.tex
\begin{scriptsize}%
\begin{table*}%
	\caption{Results on the test data set. The best 
	performance in terms of IoU per column is printed in bold, the overall greatest in blue.}%
	\begin{center} %
		\input{csv/overall_results_table_no_denoise.tex}%
	\end{center}%
	\label{table:overall_results}%
\end{table*}%
\end{scriptsize}%
\newcommand{\pathtmp}{./csv/resubmit_03/fig_confmat}
\begin{figure}[tb]
	\centering
	\subfloat[\label{fig:confmat_di}Experiment 1]{\includegraphics[height=0.4\linewidth,trim=0cm 0cm 0.6cm 0.0cm,clip] %
%	{\pathtmp/_icra_2020_resubmit_paper_02_exp01_WeatherNet_checkpoint_7000_binary.pdf}}
	{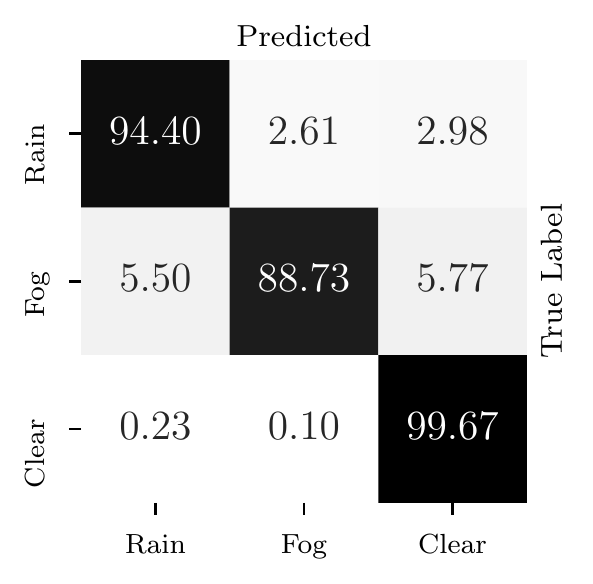}}
	\subfloat[\label{fig:confmat_di_aug}Experiment 2]{\includegraphics[height=0.4\linewidth,trim=0.835cm 0.0cm 0.60cm 0.0cm,clip] %
%	{\pathtmp/_icra_2020_resubmit_paper_02_exp02_WeatherNet_checkpoint_17000_binary.pdf}}
	{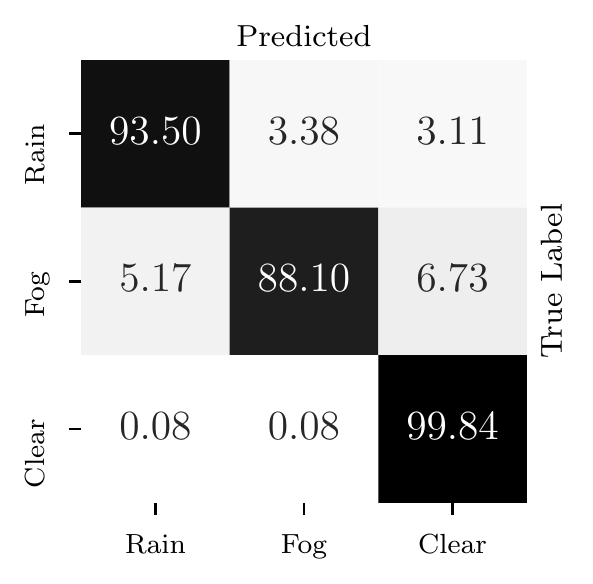}}
	\subfloat[\label{fig:confmat_di_aug_sprinkles_full}Experiment 3]{\includegraphics[height=0.4\linewidth,trim=0.835cm 0cm 0.0cm 0.0cm,clip] %
%	{\pathtmp/_icra_2020_resubmit_paper_02_exp03_WeatherNet_checkpoint_51000_binary.pdf}} %\\
	{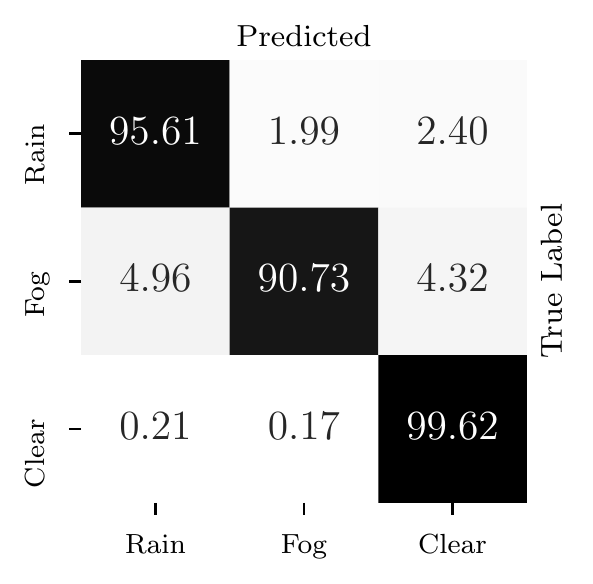}} %\\
	\centering
	\label{fig:confusion_matrix}
	\caption{Confusion matrix for \textit{WeatherNet} segmentation result.
	%Fig. \ref{fig:confmat_di} is showing the confusion matrix of experiment $1$, Fig. \ref{fig:confmat_di_aug} and Fig. \ref{fig:confmat_di_aug_sprinkles_full} of $2$ and $3$ respectively.
	}\vspace{-0.4cm}
\end{figure}%
\input{img/fig_predict_results_denoising/denoise_results_dynamic_comparison.tex}
\input{img/fig_predict_results_denoising/denoise_results_road_2.tex}
As described in section \ref{sec:method} and \ref{sec:dataset}, we obtained a large-scale data set recorded on public 
roads and in a dedicated climate chamber with different types of point-wise annotations. 
In this Section we describe several approaches to train the proposed \textit{WeatherNet} in order to maximize the 
performance and analyze the benefit of weather augmentation, especially for the generalization to natural rainfall 
recorded on roads. 
We apply the Intersection-over-Union (IoU) metric for performance evaluation, according to the 
Cityscapes Benchmark Suite \cite{Piewak2019b,Cordts2016}.
An overview of all experiments and their results is given in Table \ref{table:overall_results}. 
In order to evaluate the influence of the weather augmentation in detail, we trained the network on three different 
data subsets with and without augmentation, defined as experiment 1, 2 and 3:
\begin{enumerate}
	\item {\textbf{Chamber:} only chamber data as baseline experiment.}
	\item {\textbf{Chamber $\&$ Road:}
	Climate chamber data set and a subset of road data without any augmentation or adverse weather on roads.}
	\item {\textbf{Chamber $\&$ Road with Augmentation:} Climate chamber data set and class balanced road data set without adverse weather, but with augmentation.}
	%on road samples and climate chamber reference samples in order to evaluate the benefit of augmentations.}
%	\item {\textbf{Chamber $\&$ Road Full}\\
%	Climate chamber data set and the complete road data set for training without any augmentation.}
%	\item {\textbf{Chamber $\&$ Road Full with Augmentation}\\
%	Climate chamber data set and the complete road data set for training with augmentation on road samples and climate chamber reference 
%samples. In comparison to experiment 3, it can be evaluated whether a significant overweight of road data makes a difference.}
\end{enumerate}
Note, all evaluations are based on the test data set from experiment 2, which contains autolabeled annotations and road data without fog, rain or augmentation.
Table~\ref{table:overall_results} shows that the performance is significantly increased by using road data and the proposed weather augmentation. 
Besides validating the classes fog and rain only on chamber data, the usage of road data and the augmentation leads to an increase in the overall performance and per class IoU. This indicates that the network is able to identify weather influences in both domains and gains a general understanding of the scene.
%As the chamber recordings represent extreme conditions, in terms of size and weather, there is no significant improvement due to weather augmentation.

The results of the baseline \textit{DROR} filter indicates, that the local vicinity alone is not a proper feature to filter scatter points caused by dense water drops.
The proposed \textit{CNN} approach is outperforming \textit{DROR} by an order of magnitude.
The parameters for the \textit{DROR} are taken from \cite{Charron2018}, except for the horizontal sensor resolution which is adapted to the utilized 'VLP32C'.

Furthermore, we compare our approach to the state-of-the-art semantic segmentation models \textit{RangeNet21}, \textit{RangeNet53} \cite{Milioto2019} and \textit{LiLaNet} \cite{Piewak2019}, which provide comparable results. Consequently, we prove that the basic idea of 
\textit{CNN}-based weather segmentation and de-noising is valuable and surpasses geometrically based approaches. 
In addition, the proposed optimized \textit{WeatherNet} is mostly outperforming the other \textit{CNNs}, especially on the final experiment 3, and has a significantly lower number of trainable parameters and inference time. Thus, the network can be applied as pre-processing step. 

The confusion matrices Fig. \ref{fig:confusion_matrix} show that mostly classes rain and fog are mixed up. Since fog and rain ultimately consist of water droplets and differ only in distribution, density and size of the water droplets, this is plausible.
Moreover, lidar sensors are not designed to perceive this difference. For point cloud filtering these mix-ups are not important; a confusion is disadvantageous only with regard to classifying distinct weather conditions. 
%The very small amount of false positives, where a valid point is classified as rain or fog, is advantageous for de-noising as few valid points are removed. 
Furthermore the augmentation leads to a significant decrease in confusion between rain and fog.

\subsection{Qualitative Results on Dynamic Chamber Data}
Qualitative results on challenging dynamic scenes are presented in this section.
Whereas there is no ground truth data available due to the fact that the very same dynamic scenes cannot be recorded
under two different weather conditions. Hence, our proposed auto-labeling procedure cannot be applied.
Nevertheless, Fig. \ref{fig:denoise_results_Dynamic-G2-Fog20}, \ref{fig:denoise_results_dynamic_G6_Fog30} and \ref{fig:denoise_results_road_dense_rain} show that our approach is able to handle dynamic scenes and gives remarkable filter results.
The de-noised point cloud reveals a pedestrian and a cyclist (highlighted by black boxes) directly in front of the ego-vehicle, who almost disappear in the scatter points of the haze (Fig.\ref{fig:denoise_results_Dynamic-G2-Fog20}c, \ref{fig:denoise_results_dynamic_G6_Fog30}g).
Although the evaluated performance of \textit{RangeNet53} and \textit{WeatherNet} are comparable (Table \ref{table:overall_results}), the qualitative results show that \textit{RangeNet53} does preserve fine structures and edges of small objects (Fig. \ref{fig:denoise_results_dynamic_G6_Fog30}f), as most parts of the cyclist and pedestrian are filtered. Whereas, \textit{WeatherNet} is able to distinguish between the pedestrian/cyclist and scatter points (Fig. \ref{fig:denoise_results_dynamic_G6_Fog30}g).

A filter algorithm based on the spatial distribution of the point cloud, as shown in Fig.
\ref{fig:denoise_results_dynamic_G6_Fog30}e is not able to filter the noise in this scenario, since the fog points are similarly densely distributed as those of real objects. Nevertheless, the cyclist can be recognized slightly better, due to the ability to filter single scattering points.
Note that the DROR also filters various single points at greater distances, which are not caused by the weather.

Another benefit of our \textit{CNN}-approach is the capability of detecting the weather condition by means of lidar point clouds. 
As shown in Fig. \ref{fig:fog_clear_pointwise}, the number of scatter points caused by fog is correlated with the meteorological visibility, hence by utilizing the result of our weather segmentation, the visibility could be estimated. %
Moreover, the level of degradation of the lidar sensor could be estimated by taking the ratio of scatter to valid points into account. %
\subsection{Qualitative Results on Dynamic Road Data}
Additionally, the proposed approach is able to work on a point cloud corrupted by natural rainfall recorded on roads.
In Fig. \ref{fig:denoise_results_road_dense_rain} shows a key frame where a pedestrian is crossing the street
and multiple cars are passing by.
Despite our algorithm is only trained on climate chamber data, complemented with augmented real-world data, it shows
well performance in a real world scenario with light rain and proofs the generalization to a complete different scenario, see Fig. \ref{fig:denoise_results_road_dense_rain}.
%\subsection{Weather Detection by Lidar Sensors}
%\TODO[inline]{OPTIONAL: add results on weather estimation (visbility and/or rainfall rate) and impairment estimation}
%\subsection{Visibiity Estimation}
%\begin{figure}
%	\centering
%	\input{img/tensorboard/chambers32_fog_static_roof_valvisibility_prediction}
%	\caption{Visibility prediction based on the semantic weather segmentation. Based on the ground truth labels from the training data set, a function is fitted to estimate the visibility based on the total amount of scatter points.}
%	\label{fig:evalsummaries1}
%\end{figure}
%
%\begin{figure}[tb]
%	\centering
%	\input{img/fig_fit_x_fog_points_y_visibility/fig_fit_x_fog_points_y_visibility}
%	\caption{\DRAFT{Distribution for the multiplier coefficient for the augmentation of the measured distance. The augmentation is done for fog and rain with different densities and rainfall rates.}}
%	\label{fig:noise_distributions}
%\end{figure}

%% file: csv/overall_results_table_no_denoise.tex
\begin{scriptsize}%
\setlength{\tabcolsep}{2pt}
\begin{tabular}{l|cccc|cccc|cccc|cc}
\toprule
\multirow{3}{*}{Approach}&\multicolumn{12}{c|}{IoU in \% / Number of Samples} & Runtime$^{3)}$&Parameter$^{4)}$\\&
\multicolumn{4}{c|}{Experiment 1} &
\multicolumn{4}{c|}{Experiment 2} & 
\multicolumn{4}{c|}{Experiment 3} &
%\multicolumn{2}{c|}{Exp 1} &
%\multicolumn{2}{c|}{Exp 2} &
%\multicolumn{2}{c|}{Exp 3} &
in&in\\&
{Clear} & {Fog} & {Rain} & {Mean} & 
{Clear} & {Fog} & {Rain} & {Mean} & 
{Clear} & {Fog} & {Rain} & {Mean} &
ms & Mio \\ \toprule
%-----------------------------------------------------------number of samples
Samples Chamber$^{1)}$& 
14,386&29,777&28,637&72,800&
14,386&29,777&28,637&72,800&
14,386&29,777&28,637&72,800&
$-$ & $-$\\  
%-----------------------------------------------------------number of samples
Samples Road$^{1), 2)}$& 
$-$	  & $-$  & $-$  & $-$  & 
31,078& $-$  & $-$  & $-$  & 
34,381&34,381&34,381&103,143&
$-$ & $-$\\  \midrule
%-----------------------------------------------------------DROR
\textit{DROR} \cite{Charron2018} & 
\textbf{88.13} 	& 6.94 	& 7.37 	& 34.15 & 
88.13 	& 6.94 	& 7.37 	& 34.15 & 
88.13 	& 6.94 	& 7.37 	& 34.15 & 
100.00	& $4\mathrm{e}^{-6}$	\\ 

%\input{csv/fig_tf_summaries_resubmit_03/IoU_metrics_transpose_eval_paper.tex}
%\\\bottomrule
%\input{csv/fig_tf_summaries_resubmit_03/IoU_metrics_transpose_test_paper.tex}
\input{csv/resubmit_03/test_exp02_metrics_paper_selection.tex}
\\
\bottomrule
\multicolumn{5}{r}{$^{1)}$ the column \textit{Mean} states the total number the samples}&
\multicolumn{4}{r}{$^{2)}$ fog and rain are augmented samples}&
\multicolumn{3}{r}{$^{3)}$ on GeForce GTX 1080 Ti}&
\multicolumn{3}{r}{$^{4)}$ number of trainable parameters}
\end{tabular}\vspace{-0.75cm}
\end{scriptsize}%
%lspci -nn | grep '\[03'
%1a:00.0 VGA compatible controller [0300]: NVIDIA Corporation GP102 [GeForce GTX 1080 Ti] [10de:1b06] (rev a1)
%68:00.0 VGA compatible controller [0300]: NVIDIA Corporation GP102 [GeForce GTX 1080 Ti] [10de:1b06] (rev a1)

%% file: csv/resubmit_03/test_exp02_metrics_paper_selection.tex
%clear&fog&rain&mean
%\textit{-RangeNet53-v2-ckp-45000}&%---------------------RangeNet53-v2-ckp-45000
\textit{RangeNet53} \cite{Milioto2019}&%--------------------RangeNet53
74.73&77.32&\textcolor{maxcolor}{91.22}&81.09
&87.75&\textcolor{maxcolor}{86.46}&\textcolor{maxcolor}{94.23}&\textcolor{maxcolor}{89.48}
&86.50&87.19&\textcolor{maxcolor}{94.36}&89.35
&51.90&66.17	%RT and params
\\
%\textit{-RangeNet21-v2-ckp-20000}&%---------------------RangeNet21-v2-ckp-20000
\textit{RangeNet21} \cite{Milioto2019}&%--------------------RangeNet53
71.53&71.40&86.13&76.35
&86.71&80.90&87.01&84.87
&85.10&79.94&85.35&83.46
&33.83&38.50 	%RT and params
\\
%\textit{-LiLaNet-ckp-95000}&%---------------------LiLaNet-ckp-95000
\textit{LiLaNet} \cite{Piewak2019b}&%--------------------LiLaNet
82.72&\textbf{79.57}&88.16&\textbf{83.48}
&\textcolor{maxcolor}{91.60}&84.96&88.62&\textbf{88.39}
&\textcolor{maxcolor}{93.85}&\textbf{88.74}&90.82&\textcolor{maxcolor}{91.14}
&91.93&7.84		%RT and params
\\
%\textit{-WeatherNet-ckp-50000}&%---------------------WeatherNet-ckp-50000
\textit{WeatherNet} \textit{[Ours]}&%--------------------WeatherNet
\textcolor{maxcolor}{91.65}&\textcolor{maxcolor}{86.40}&\textbf{89.29}&\textcolor{maxcolor}{89.11}
&\textbf{90.89}&\textbf{85.15}&\textbf{88.84}&88.29
&\textbf{93.35}&\textcolor{maxcolor}{88.81}&\textbf{90.92}&\textbf{91.03}
&34.45&1.53		%RT and params

%% file: img/fig_predict_results_denoising/denoise_results_dynamic_comparison.tex
%\tikzstyle{txt}=[fill=none,inner sep=1pt, text centered,minimum height=1.0em, font=\footnotesize]
%\tikzstyle{box}=[fill=white,inner sep=1pt,text width=7.42cm, text centered,minimum height=1.0em]
%\tikzstyle{txt2}=[draw,fill=white,inner sep=1pt,text centered,minimum width=1.0em, text height=1.5cm]
\newcommand{\mylegend}[4]{
\node (#1) [fill=white, right=1.0cm of #2, font=\footnotesize]  {#3};
\node (#1box) [rectangle, left=0.05cm of #1, minimum width=0.5cm, color=black, fill=#4, draw] {};
}
\newcommand{\includeTmp}[3]{
\begin{tikzpicture}
	\node[anchor=south west,inner sep=0, shift={(0,0)}] (image) at (0,0) {
	\includegraphics[width=0.128\linewidth,trim=11.1cm 0.750cm 11.1cm 3.50cm,clip]{#1}
	};
	\begin{scope}[x={(image.south east)},y={(image.north west)}]
		 \node [anchor=north east, draw, fill=none,thick, text width=1.25em,text centered, minimum height=2.50em] (box) at (0.27,0.60) {};	
		 \node [anchor=north east, fill=white, text centered, above right = -0.015cm and -0.090cm of box]() {\footnotesize Cyclist};	
	 \end{scope}	 
	\node[anchor=south west,inner sep=0, shift={(0,2.5)}] (image2) at (0,0) {
	\includegraphics[width=0.128\linewidth,trim=11.1cm 0.750cm 11.1cm 3.50cm,clip]{#3}};
	\begin{scope}[x={(image2.south east)},y={(image2.north west)}]
	 \node [anchor=north east, draw, fill=none,thick, text width=1.1em,text centered, minimum height=2.450em] (box2) at (0.4350,0.590) {};
	\node [anchor=north east, fill=white, text centered, above = -0.015cm of box2]() {\footnotesize Ped.};
 \end{scope}
\end{tikzpicture}
}%
\begin{figure*}
\centering
\begin{tikzpicture}
 \node[anchor=south west,inner sep=0, shift={(0,0)}] (image) at (0,0) %
 {
	% tim left bottom right top
	\subfloat[\label{fig:denoise_results_dynamic_G6_Fog30a}Raw Data]%
	{\includeTmp{\resultFolder\weatherNetinUse\pathCycC_gt.png}{}
	{\resultFolder\weatherNetinUse\pathPedC_gt.png}} %
%	%
	\subfloat[\label{fig:denoise_results_dynamic_comparison_seg_DROR}\textit{DROR}]%
	{\includeTmp{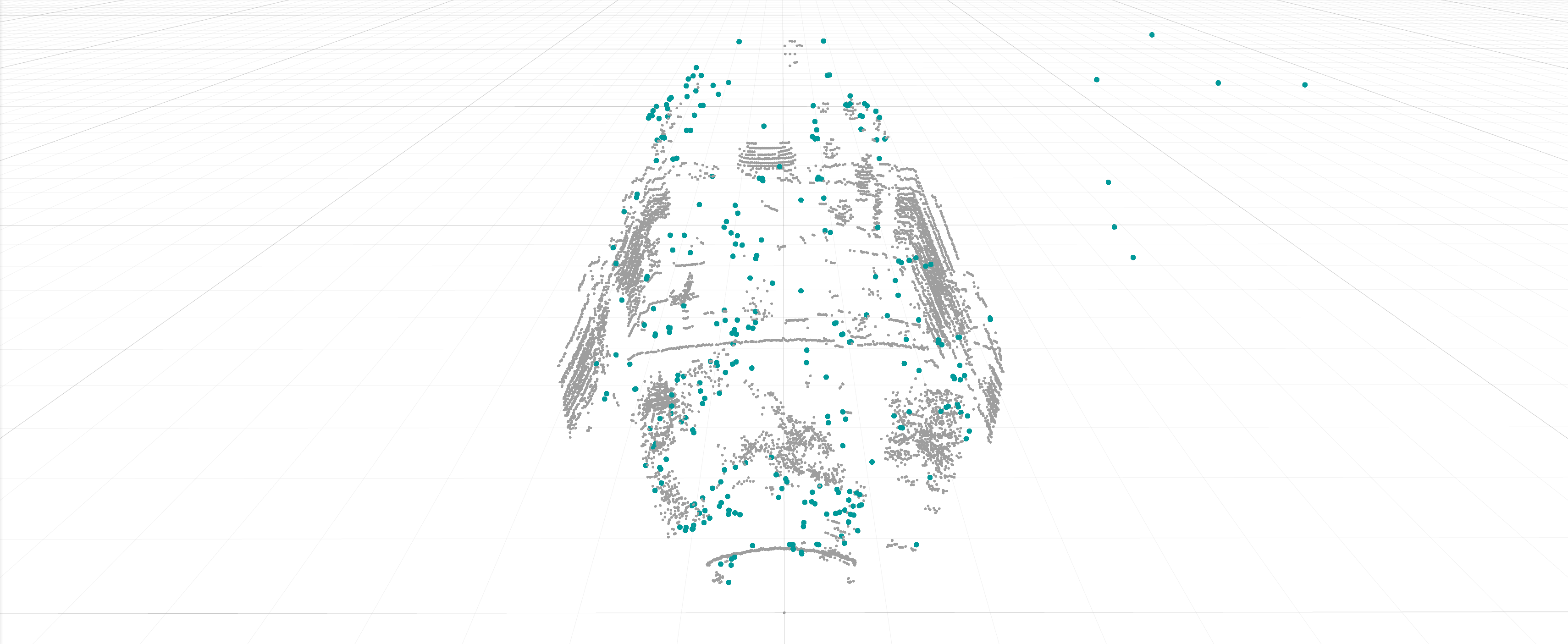}{}%
	{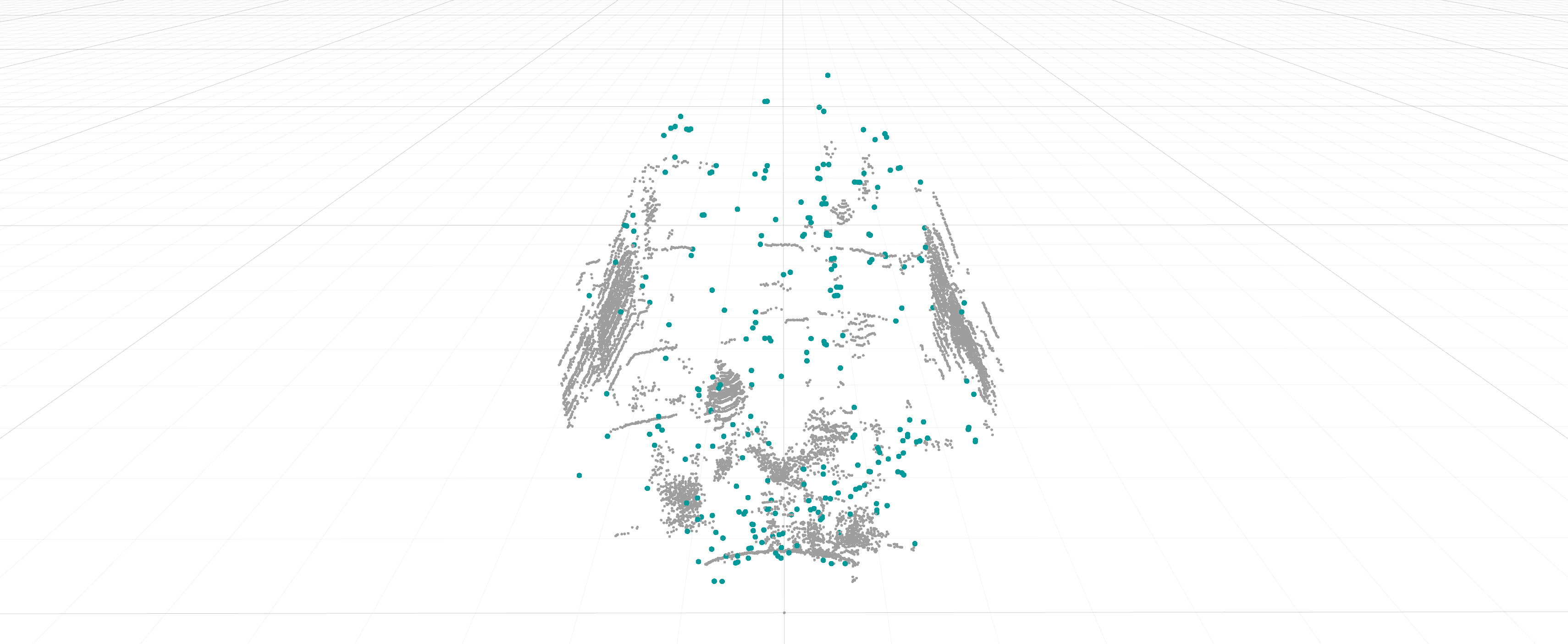}} %
%	%
	\subfloat[\label{fig:denoise_results_dynamic_comparison_seg_RN}\textit{RangeNet53}]%
	{\includeTmp{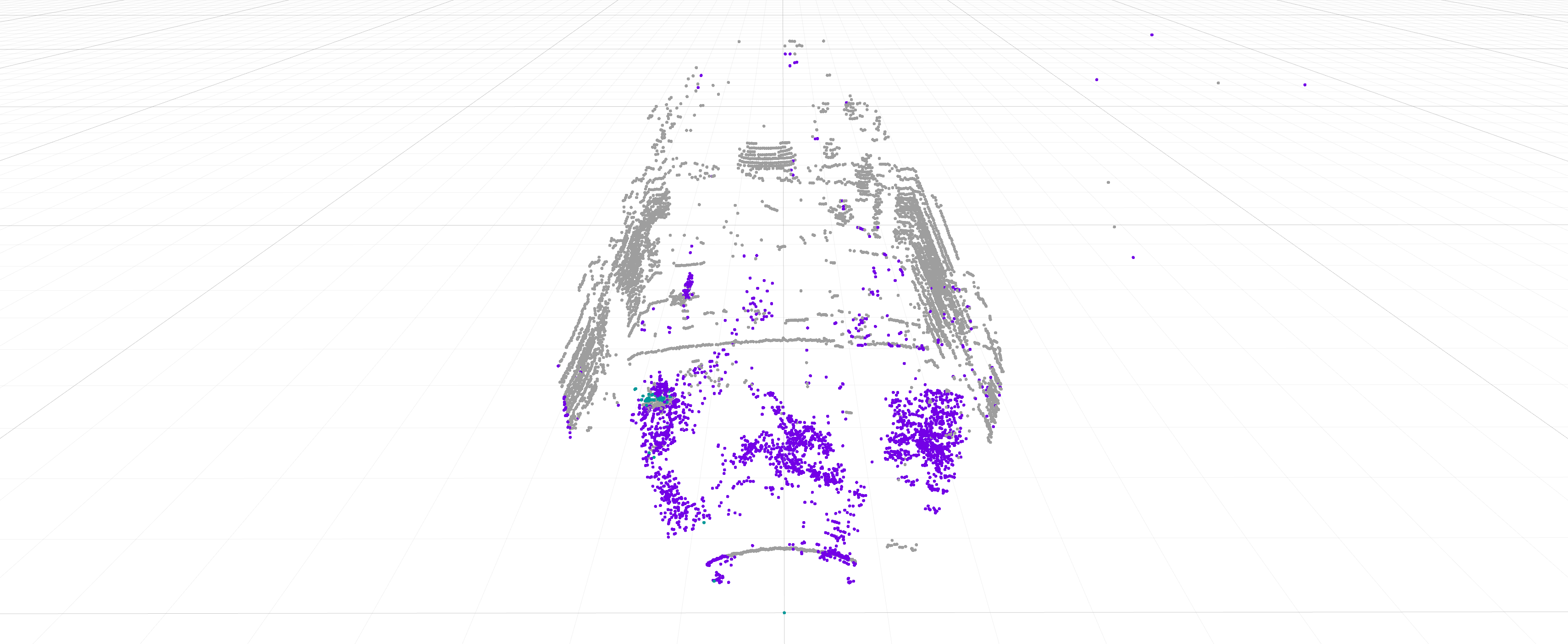}{} %
	{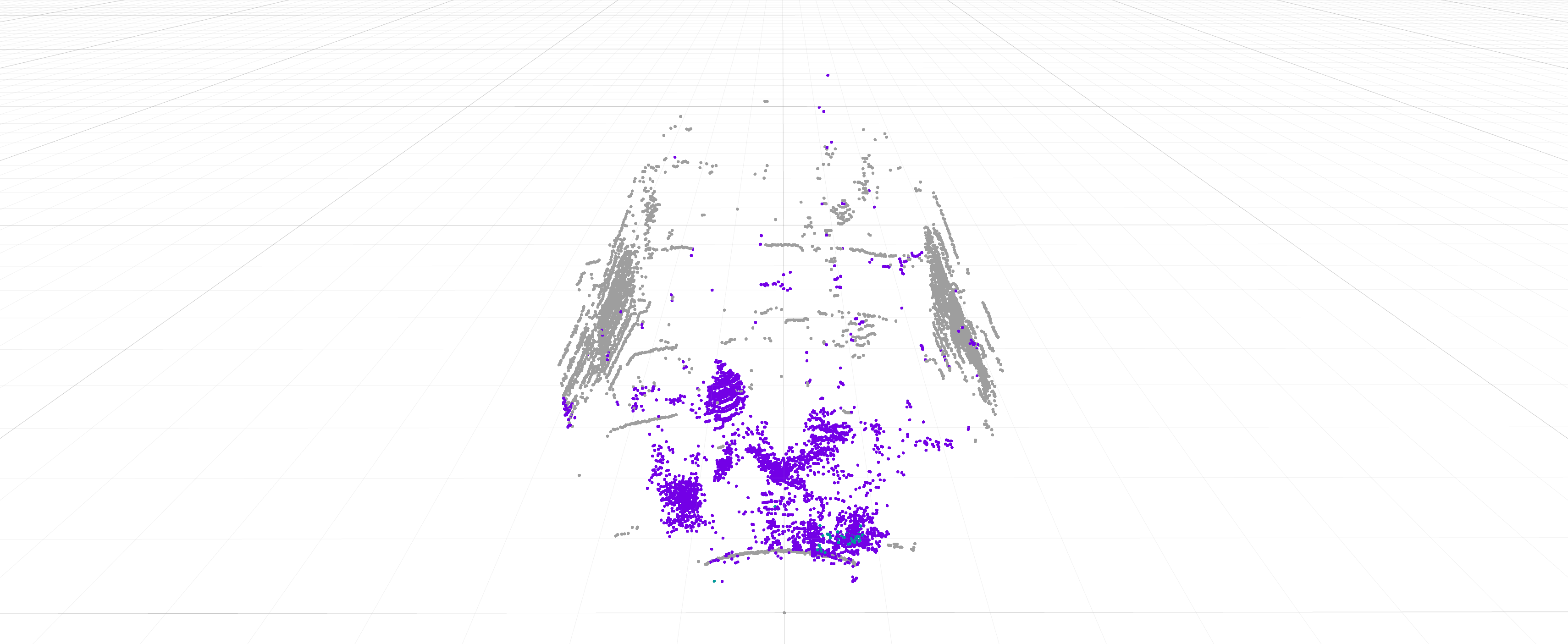}} %
%	%
	\subfloat[\label{fig:denoise_results_dynamic_comparison_seg_WN}\textit{WeatherNet}]%
	{\includeTmp{\resultFolder\weatherNetinUse\pathCycC_prediction_solid.png}{} %
	{\resultFolder\weatherNetinUse\pathPedC_prediction_solid.png}}%
	\subfloat[\label{fig:denoise_results_dynamic_comparison_DROR}\textit{DROR}]%
	{\includeTmp{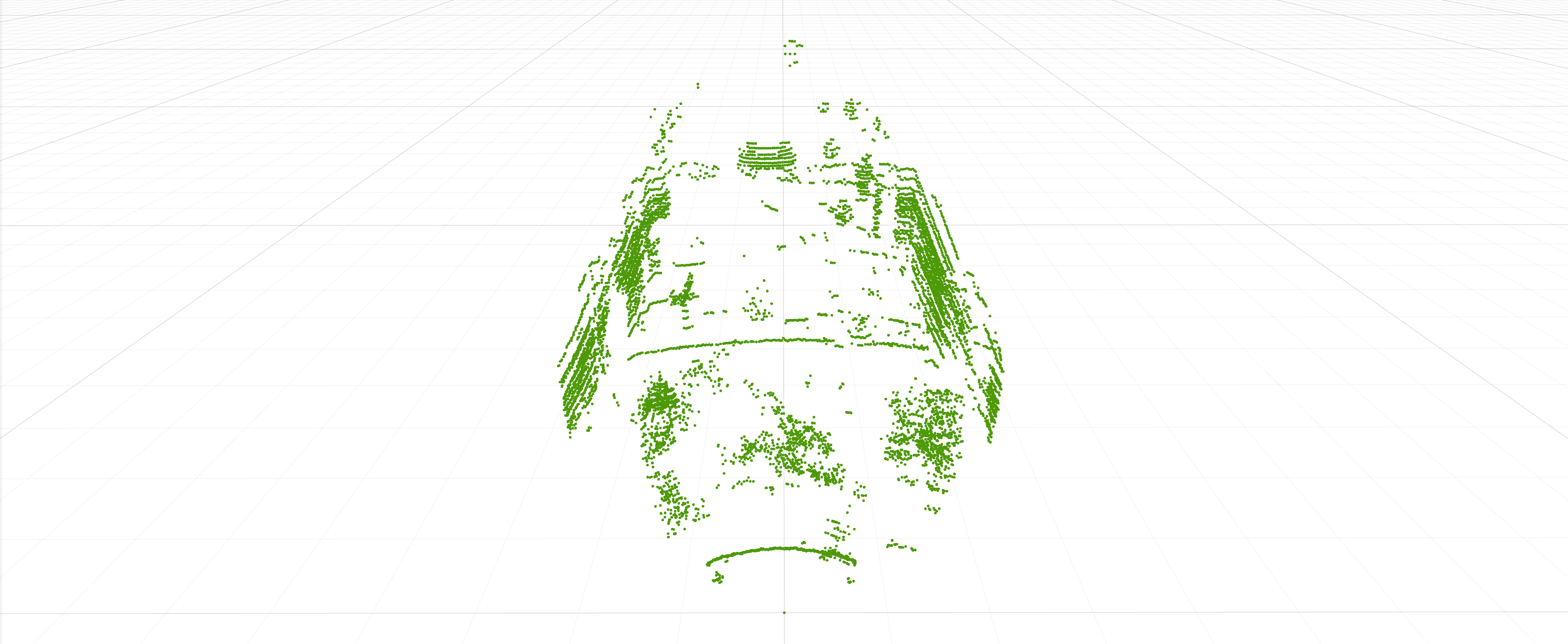}{} %
	{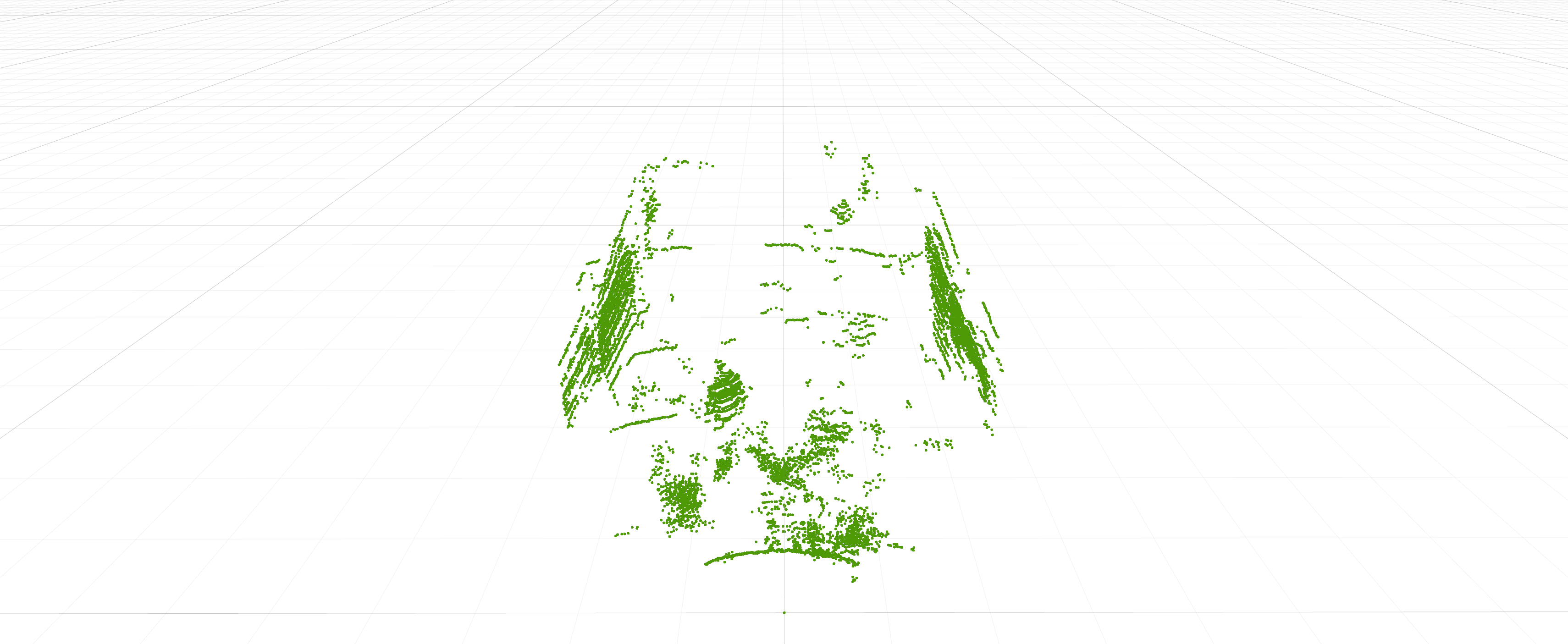}} %
	\subfloat[\label{fig:denoise_results_dynamic_comparison_RN}\textit{RangeNet53}]%
	{\includeTmp{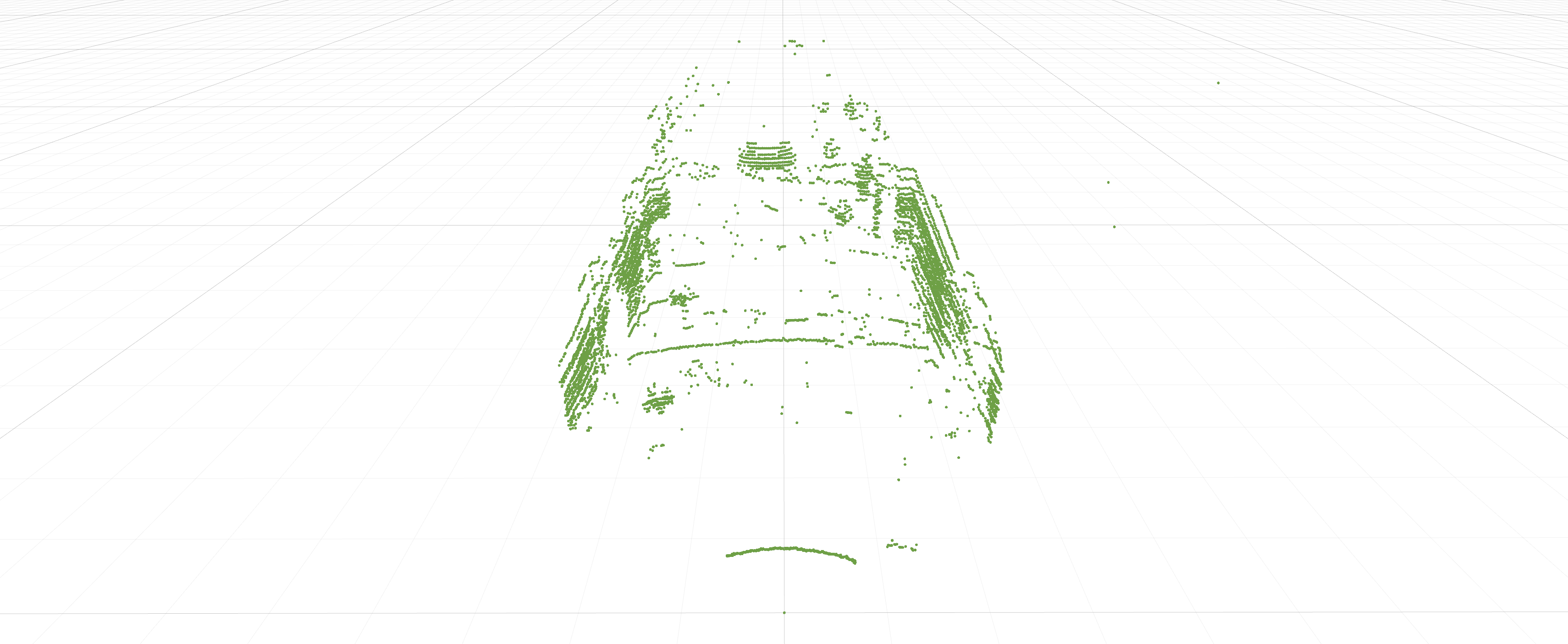}{}%
	{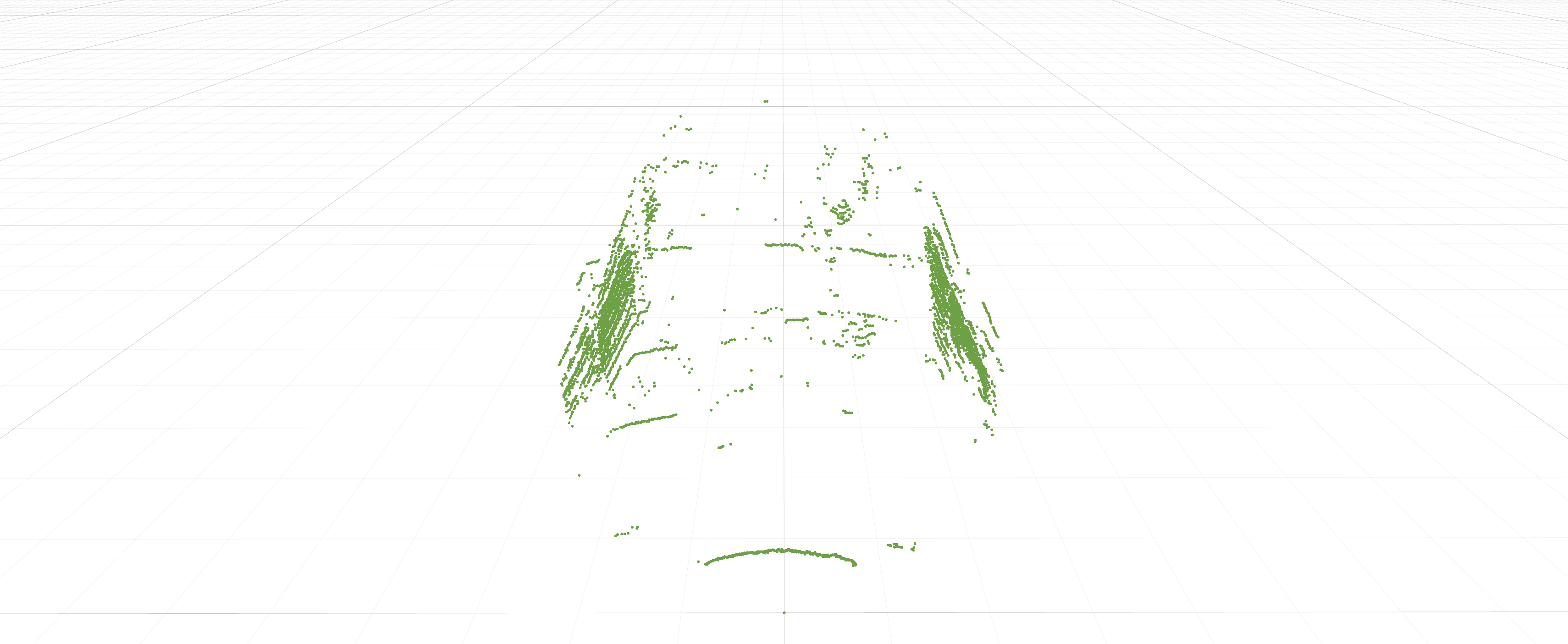}}
	\subfloat[\label{fig:denoise_results_dynamic_comparison_WN}\textit{WeatherNet}]%
	{\includeTmp{\resultFolder\weatherNetinUse\pathCycC_denoised.png}{}%
	{\resultFolder\weatherNetinUse\pathPedC_denoised.png}}
};
\begin{scope}[x={(image.south east)},y={(image.north west)}]

% legend input
%text width=2.45cm % 1 col
%text width=7.5cm, % 3 cols
\node [box, fill=none,minimum height=14.5em] 	(box1) at (0.075,0.50) {};
\node [box, text width=2.45cm, above = -0.2cm of box1] 				() {};
\node [txt, above = -0.2cm of box1] 								(txt1) {Input};
\node (l1box) [rectangle, left=0.05cm of txt1, minimum width=0.25cm, color=white, fill=colorInput, draw] {};
% legend segmentation
\node [box, fill=none, text width=7.42cm, minimum height=14.5em,  right = 0.0cm of box1] 	(box2) {};
\node [box,   above = -0.2cm of box2] 			() {};
\node [txt, above = -0.2cm of box2] 									(txt2) {Fog};
\node [txt, above right = -0.2cm and -2.95cm of box2] 					(txt3) {Rain};
\node [txt, above left = -0.2cm and -3.0cm of box2] 					(txt4) {Valid};
\node (l2box) [rectangle, left=0.05cm of txt2, minimum width=0.25cm, color=white, fill=colorFog, draw] {};
\node (l3box) [rectangle, left=0.05cm of txt3, minimum width=0.25cm, color=white, fill=colorRain, draw] {};
\node (l4box) [rectangle, left=0.05cm of txt4, minimum width=0.25cm, color=white, fill=colorValid, draw] {};
% legend de noised
\node [box, fill=none, text width=7.42cm, minimum height=14.5em, right = 0.0cm of box2] 	(box3) {};
\node [box,  above = -0.2cm of box3] 									() {};
\node [txt, above = -0.2cm of box3] 									(txt5) {De-Noised};
\node (l5box) [rectangle, left=0.05cm of txt5, minimum width=0.25cm,color=white,fill=colorDenoised, draw] {};

%\node [txt2] at (0.0,0.650) {{\footnotesize \rotatebox[origin=c]{90}{Pedestrian}}};
%\node [txt2] at (0.0,0.250) {{\footnotesize \rotatebox[origin=c]{90}{Cyclist}}};
\end{scope}
\end{tikzpicture}
	
	\caption{De-noising results shown on a snapshot with two dynamic objects in dense fog at $30\,$m visibility. The color coding is similar to Fig. \ref{fig:denoise_results_Dynamic-G2-Fog20}. For the baseline \textit{DROR} filter all points are  colored as rain points, since no distinction between rain and fog is possible. In addition the segmentation and de-noising results from \textit{RangeNet53} and \textit{WeatherNet} are given. The cyclist and pedestrian, which are barely recognizable in the scene, are highlighted in a black box. Note that the pedestrian and cyclist remains after filtering by \textit{WeatherNet} while discarding the fog clutter.}\vspace{-0.2cm}
	\label{fig:denoise_results_dynamic_G6_Fog30}
\end{figure*}
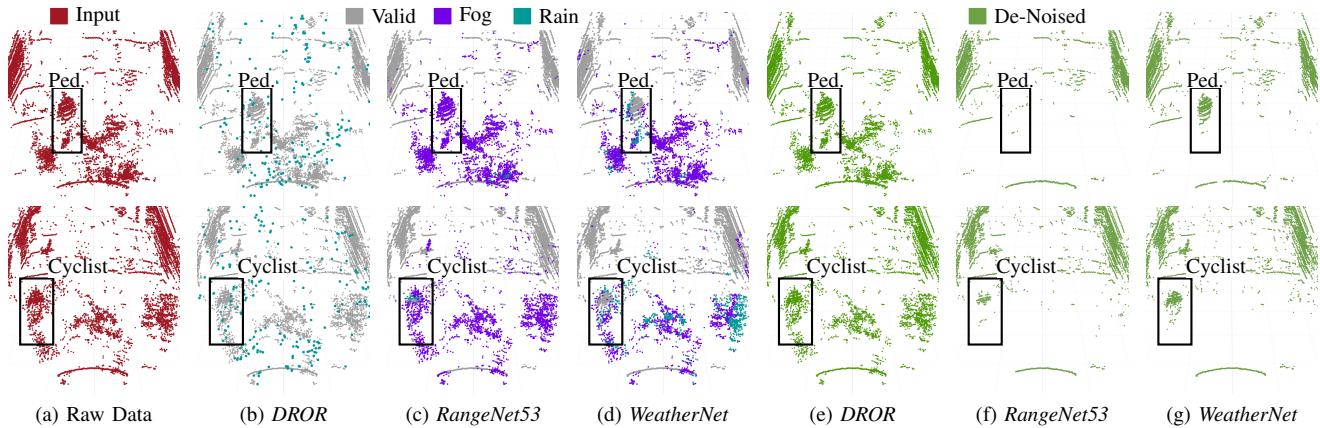

%% file: img/fig_predict_results_denoising/denoise_results_road_2.tex
\renewcommand{\includeTmp}[3]{
\begin{tikzpicture}
 \node[anchor=south west,inner sep=0, shift={(0,0)}] (image) at (0,0) {\includegraphics[width=0.49\linewidth, %
%trim=left bottom right top, clip]
 trim=7.0cm 2.00cm 4.00cm 2.0cm,clip]{#1}};
%trim=7.0cm 2.0cm 4.00cm 0.0cm,clip]{#1}};
 \begin{scope}[x={(image.south east)},y={(image.north west)}]
  #3
 \node [anchor=north east,text centered,black,fill=white,opacity=1.0,text opacity=1](text) at (1.01,0.1){\footnotesize #2};
 \node [anchor=north east, draw, fill=none,thick, text width=1.0em,minimum height=2.250em]%
 (box2) at (0.45,0.550) {};
% \node [anchor=north east, fill=white, text centered,above left = -0.035cm and -0.080cm of box2](){\footnotesize Ped.};
\node [anchor=north east, draw,fill=none,thick, text width=2.0em,minimum height=1.75em]%
(box2) at (0.490,0.78) {};
%\node [anchor=north east, fill=white, text centered, above=-0.01cm of box2](){\footnotesize Car};
\node [anchor=north east, draw,fill=none,thick, text width=2.6em,minimum height=1.75em]%
(box2) at (0.289,0.74) {};
%\node [anchor=north east, fill=white, text centered, above=-0.01cm of box2](){\footnotesize Car};
 \end{scope}
\end{tikzpicture}
}%
\begin{figure*}%
\includeTmp{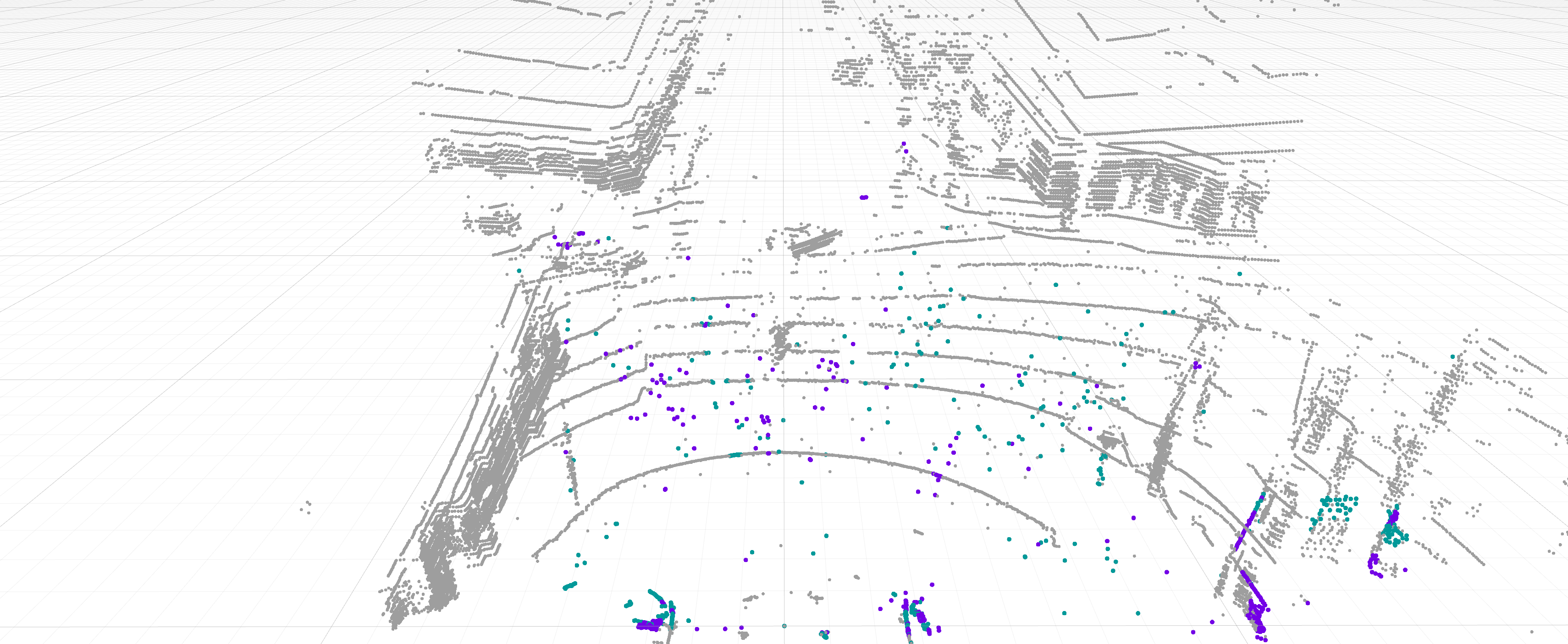}{Exp 2: No Augmentation}
{
	\node [anchor=north east] at (1.0150,1.040) {\includegraphics[width=0.175\linewidth,trim=1.0cm 0.00cm 7.5cm 0.0cm,clip]{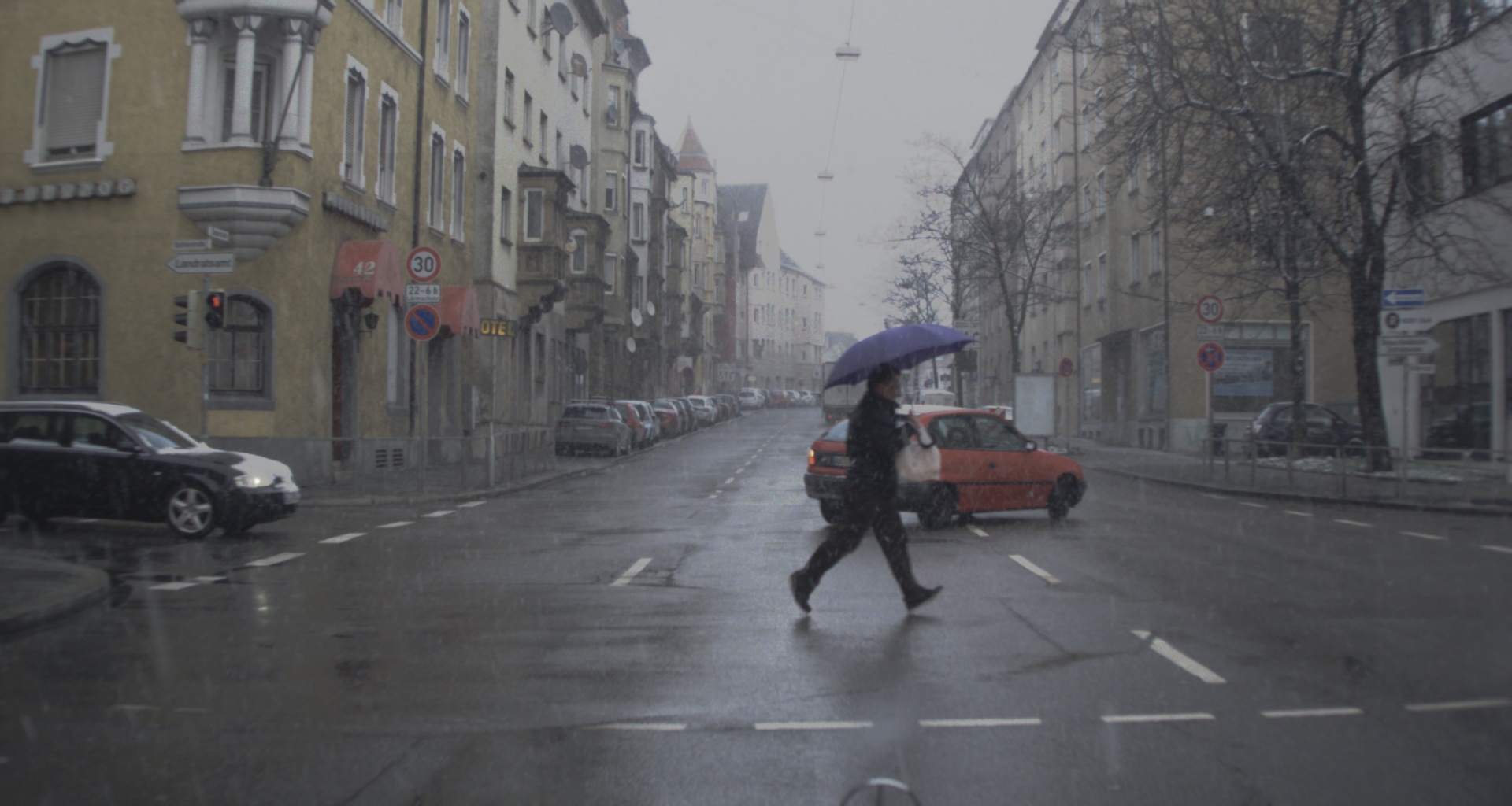}};
}%
\includeTmp{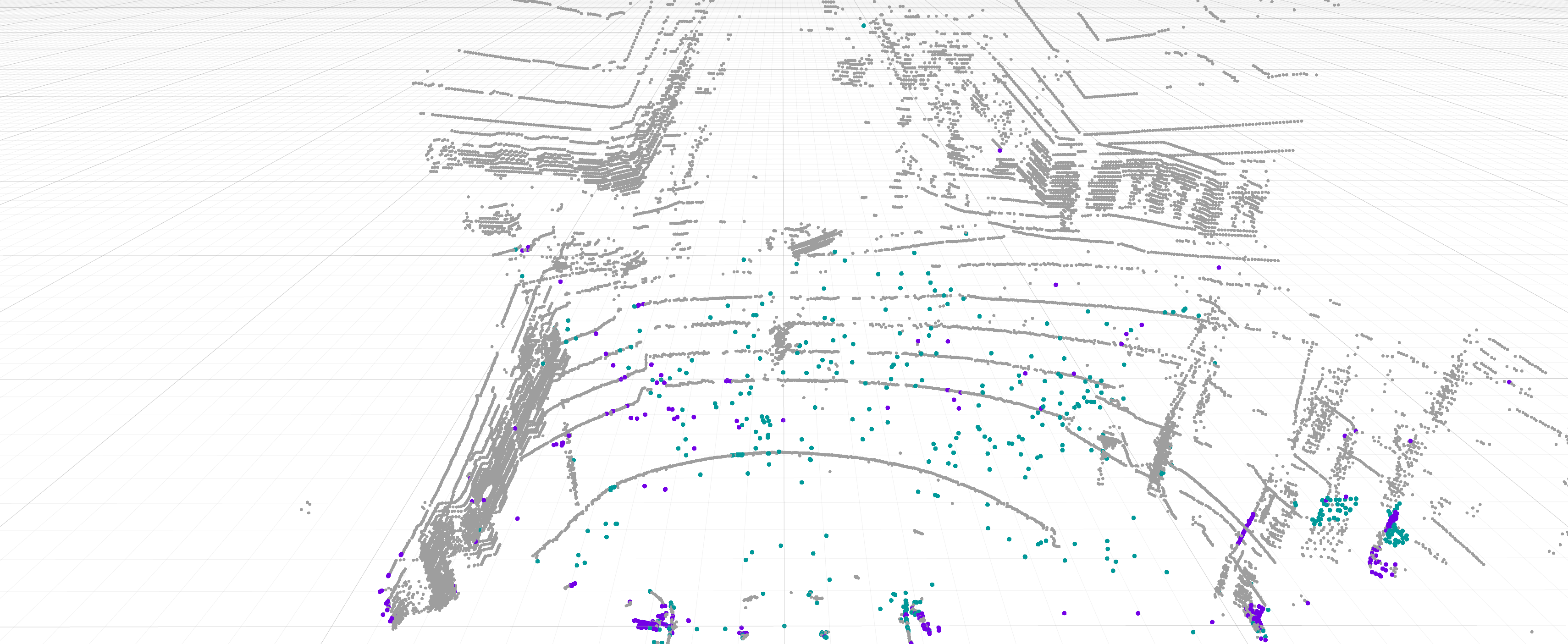}{Exp 3: With Augmentation}
{ 	
	\node [anchor=north east] at (1.0150,1.040) {\includegraphics[width=0.175\linewidth,trim=1.0cm 0.00cm 7.5cm 0.0cm,clip]{./csv/resubmit_03/2018-01-17_143024_LidarImage_000000920_cam_frame_000001827.jpg}};
}%
\caption{\textit{WeatherNet} segmentation results for road data recorded under light rainfall. The result shows, that the approach is able to generalize the noise pattern and predicts well results for recordings during natural rainfall. Note that the training data set neither contains natural rain nor fog scenes on roads. The color coding is similar to Fig. \ref{fig:denoise_results_Dynamic-G2-Fog20} and the objects shown in the camera image are highlighted with a black box. The training with augmentation (right) leads to a better segmentation result in terms of number of detected raindrops and less false negatives for object detection (e.g. left car).}\vspace{-0.4cm}
\label{fig:denoise_results_road_dense_rain}
\end{figure*}

%% file: tex/conclusion.tex
We presented a \textit{CNN}-based approach for point cloud weather segmentation as an essential pre-processing step for
lidar-based environment perception to distinguish between scatter points from adverse weather and valid points from
solid objects. As opposed to previous approaches that analyze the statistics of the local spatial vicinity of
individual points, we opted for a learning-based approach that involves a global understanding 
%and consideration 
of a
traffic scene as a whole to estimate the validity of point-level measurements. The issue of requiring annotated
ground truth data for our approach is significantly alleviated by our proposed data augmentation strategy. Our
quantitative and qualitative results demonstrate the superior performance of our \textit{CNN}-based approach over state-of-the-art,
while being very efficient at the same time. 
For more qualitative results on dynamic scenes we refer to our data set page (link in the abstract) and the supplementary video.%

%% file: tex/acknowledgment.tex
This work was done in cooperation with the Dense Project, contract no. 692449, of the European Union under the H2020 ECSEL programme.
%We thank  and Mario Bijelic for help recording the dataset.
We thank Mario Bijelic and Tobias Gruber for enabling this work by recording the data set.